\documentclass{article}

\usepackage[preprint,nonatbib]{neurips_2022}

\usepackage[utf8]{inputenc} 
\usepackage[T1]{fontenc}    
\usepackage{hyperref}       
\usepackage[hyperpageref]{backref}

\usepackage{url}            
\usepackage{booktabs}       
\usepackage{amsfonts}       
\usepackage{nicefrac}       
\usepackage{microtype}      
\usepackage{xcolor}         
\usepackage{lipsum}
\usepackage{amsmath}

\usepackage{enumitem}
\usepackage{graphicx}
\usepackage{adjustbox}

\newcommand{\lrf}[1]{\left\{ {#1} \right\}}

\newcommand{\R}{\mathbb{R}}
\newcommand{\X}{\mathbb{X}}
\newcommand{\Y}{\mathbb{Y}}

\newcommand\extrafootertext[1]{%
    \bgroup
    \renewcommand\thefootnote{\fnsymbol{footnote}}%
    \renewcommand\thempfootnote{\fnsymbol{mpfootnote}}%
    \footnotetext[0]{#1}%
    \egroup
}

\title{Revisiting Pretraining Objectives for \\ Tabular Deep Learning}

\author{%
Ivan Rubachev$^{\alpha,\beta}$ \quad Artem Alekberov$^{\alpha,\beta}$ \quad  Yury Gorishniy \quad  Artem Babenko$^{\alpha}$ \\
$^\alpha$Yandex \qquad $^\beta$HSE University
}

\begin{document}

\setlength{\tabcolsep}{2pt}

\extrafootertext{Correspondence to \href{mailto:irubachev@gmail.com}{\texttt{irubachev@gmail.com}}}
\extrafootertext{Code available at \href{https://github.com/puhsu/tabular-dl-pretrain-objectives}{\texttt{github.com/puhsu/tabular-dl-pretrain-objectives}}}

\maketitle

\begin{abstract}
Recent deep learning models for tabular data currently compete with the traditional ML models based on decision trees (GBDT). Unlike GBDT, deep models can additionally benefit from pretraining, which is a workhorse of DL for vision and NLP.
For tabular problems, several pretraining methods were proposed, but it is not entirely clear if pretraining provides consistent noticeable improvements and what method should be used, since the methods are often not compared to each other or comparison is limited to the simplest MLP architectures.

In this work, we aim to identify the best practices to pretrain tabular DL models that can be universally applied to different datasets and architectures.
Among our findings, we show that using the object target labels during the pretraining stage is beneficial for the downstream performance and advocate several target-aware pretraining objectives.
Overall, our experiments demonstrate that properly performed pretraining significantly increases the performance of tabular DL models, which often leads to their superiority over GBDTs.

\end{abstract}

\section{Introduction}

Tabular problems are ubiquitous in industrial ML applications, which include data described by a set of heterogeneous features, such as learning-to-rank, click-through rate prediction, credit scoring, and many others. Despite the current dominance of deep learning models in the ML literature, for tabular problems, the ``old-school'' decision tree ensembles (e.g., GBDT) are often the top choice for practitioners. Only recently, several works have proposed the deep models that challenge the supremacy of GBDT in the tabular domain \cite{tabnet,revisiting,saint,gorishniy2022embeddings} and suggest that the question ``tabular DL or GBDT'' is yet to be answered.

An important advantage of deep models over GBDT is that they can potentially achieve higher performance via pretraining their parameters with a properly designed objective.
These pretrained parameters, then, serve as a better than random initialization for subsequent finetuning for downstream tasks.
For computer vision and NLP domains, pretraining is a de facto standard and is shown to be necessary for the state-of-the-art performance \cite{he2021masked, BERT}. For tabular problems, however, such a consensus is yet to be achieved as well as the best practices of tabular pretraining are to be established. In particular, pretraining for tabular problems is typically performed directly on the downstream target datasets, unlike pretraining in vision or NLP problems, for which huge ``extra'' data is available on the Internet.
While a large number of prior works addresses the pretraining of tabular DL models \cite{vime, scarf, subtab, darabi2021contrastive}, it is challenging to make reliable conclusions about pretraining efficacy in tabular DL from the literature since experimental setups vary significantly. Some evaluation protocols assume the unlabeled data is abundant but use a small subset of labels from each dataset during finetuning for evaluation -- demonstrating pretraining efficacy, but somewhat limiting the performance of supervised baselines.

By contrast, in our work, we focus on the setup with fully labeled tabular datasets to understand if pretraining helps tabular DL in a fully supervised setting and compare pretraining methods to the strong supervised baselines. To this end, we perform a systematic experimental evaluation of several pretraining objectives, identify the superior ones, and describe the practical details of how to perform tabular pretraining optimally. Our main findings, which are important for practitioners, are summarized below:

\begin{itemize}

    \item Pretraining provides substantial gains over well-tuned supervised baselines in the fully supervised setup.

    \item Simple self-prediction based pretraining objectives are comparable to the objective based on contrastive learning. To the best of our knowledge, this was not reported before in tabular DL.

    \item The object labels can be exploited for more effective pretraining. In particular, we describe several ``target-aware'' objectives and demonstrate their superiority over their ``unsupervised'' counterparts.

    \item The pretraining provides the most noticeable improvements for the vanilla MLP architecture. In particular, their performance after pretraining becomes comparable to the state-of-the-art models trained from scratch, which is important for practitioners, who are interested in simple and efficient solutions.

    \item The ensembling of pretrained models is beneficial. It indicates that the pretraining stage does not significantly decrease the diversity of the models, despite the fact that all the models are initialized by the same set of parameters.

\end{itemize}

Overall, our work provides a set of recipes for practitioners interested in tabular pretraining, which results in higher performance for most of the tasks. The code of our experiments is available online.

\section{Related Work}

Here we briefly review the lines of research that are relevant to our study.

\textbf{Status Quo in tabular deep learning.}
A plethora of recent works have proposed a large number of deep models for tabular data \cite{snn, node, tabnet, autoint, dcn, grownet, tel, tabtransformer, saint, revisiting, npt}. Several systematic studies, however, reveal that these models typically do not consistently outperform the decision tree ensembles, such as GBDT (Gradient Boosting Decision Tree) \cite{xgboost, catboost, lightgbm}, which are typically the top-choice in various ML competitions \cite{revisiting, dl_is_not}. Additionally, several works have shown that the existing sophisticated architectures are not consistently superior to properly tuned simple models, such as MLP and ResNet \cite{revisiting, cocktails}. Finally, the recent work \cite{gorishniy2022embeddings} has highlighted that the appropriate embeddings of numerical features in the high-dimensional space are universally beneficial for different architectures. In our work, we experiment with pretraining of both traditional MLP-like models and advanced embedding-based models proposed in \cite{gorishniy2022embeddings}.

\textbf{Pretraining in deep learning.} For domains with structured data, like natural images or texts, pretraining is currently an established stage in the typical pipelines, which leads to higher general performance and better model robustness \cite{he2021masked,BERT}.
Pretraining with the auto-encoding objective was also previously studied as a regularization strategy helping in the optimization process \cite{unsup-pretraining-help, pretrain-on-target} without large scale pretraining datasets. During the last years, several families of successful pretraining methods have been developed. An impactful line of research on pretraining is based on the paradigm of contrastive learning, which effectively enforces the invariance of the learned representations to the human-specified augmentations \cite{chen2020simple, he2020momentum}.  Another line of methods exploits the idea of self-prediction, i.e., these methods require the model to predict certain parts of the input given the remaining parts \cite{he2021masked, BERT}. In the vision community, the self-prediction based methods are shown to be superior to the methods that use contrastive learning objectives \cite{he2021masked}. In our experiments, we demonstrate that self-prediction based objectives are comparable to the contrastive learning ones on tabular data, while being much simpler.

\textbf{Pretraining for the tabular domain.} Numerous pretraining methods were recently proposed in several recent works on tabular DL \cite{tabnet, vime, darabi2021contrastive, subtab, saint, npt}. However, most of these works do not focus on the pretraining objective per se and typically introduce it as a component of their tabular DL pipeline. Moreover, the experimental setup varies significantly between methods. Therefore, it is difficult to extract conclusive evidence about pretraining effectiveness from the literature. To the best of our knowledge, there is only one systematic study on the tabular pretraining \cite{scarf}, but its experimental evaluation is performed only with the simplest MLP models, and we found that
the superiority of the contrastive pretraining, reported in \cite{scarf}, does not hold for tuned models in our setup, where contrastive objective is comparable to the simpler self-prediction objectives.

\section{Revisiting pretraining objectives}
\label{sec:revisiting_objectives}

In this section, we evaluate the typical pretraining objectives under the unified experimental setup on the number of datasets from the literature on tabular DL.
Our goal is to answer whether pretraining generally provides significant improvements in downstream task performance over tuned models trained from scratch and to identify the pretraining objectives that lead to the best downstream task performance.

\subsection{Experimental setup}
\label{sec:experimental_setup}

We mostly follow the experimental setup from \cite{gorishniy2021revisiting} and describe its main details here for completeness.

\textbf{Notation.} Each tabular dataset is represented by a set of pairs $\lrf{(x_i, y_i)}_{i=1}^n$, where $x_i = (x_{i}^{1}, \ldots, x_{i}^{m}) \in \X$ are the objects features (both numerical and categorical) and $y_i \in \Y$ is the target variable. The downstream task is either regression $\Y = \R$ or classification $\Y = \lrf{1,\ldots,k}$. Each model has the backbone $f(x | \theta)$ that is followed by two separate heads: a pretraining head $h(z | \mu)$ and a downstream task head $g(z | \lambda)$, with learnable parameters $\theta, \mu, \lambda$ respectively, and $z = f(x | \theta)$ denotes the output of the backbone for an input object $x$.

\textbf{Datasets.} We evaluate the pretraining methods on a curated set of eleven middle to large scale datasets used in prior literature on tabular deep learning. The benchmark is biased towards tasks, where tuned MLP models were shown to be inferior to GBDT \cite{gorishniy2021revisiting} since we aim to understand if pretraining can help the deep models to beat the ``shallow'' ones. The datasets represent a diverse set of tabular data problems with classification and regression targets. The main dataset properties are summarized in \autoref{tab:dataset_info}.

\begin{table}[h!]
    \centering
    \caption{Datasets used for the experiments}
    {\small \begin{tabular}{lcccccccc}
\toprule
Abbr & Name & \# Train & \# Validation & \# Test & \# Num & \# Cat & Task type & Batch size \\
\midrule
GE & Gesture Phase & $6318$ & $1580$ & $1975$ & $32$ & $0$ & Multiclass & 128 \\
CH & Churn Modelling & $6400$ & $1600$ & $2000$ & $10$ & $1$ & Binclass & 128 \\
CA & California Housing & $13209$ & $3303$ & $4128$ & $8$ & $0$ & Regression & 128 \\
HO & House 16H & $14581$ & $3646$ & $4557$ & $16$ & $0$ & Regression & 128 \\
AD & Adult ROC & $26048$ & $6513$ & $16281$ & $6$ & $8$ & Binclass & 256 \\
OT & Otto Group Products LogLoss & $39601$ & $9901$ & $12376$ & $93$ & $0$ & Multiclass & 256 \\
HI & Higgs Small & $62751$ & $15688$ & $19610$ & $28$ & $0$ & Binclass & 512 \\
FB & Facebook Comments Volume & $157638$ & $19722$ & $19720$ & $50$ & $1$ & Regression & 512 \\
WE & Shifts Weather (subset) & $296554$ & $47373$ & $53172$ & $123$ & $0$ & Regression & 1024 \\
CO & Covertype & $371847$ & $92962$ & $116203$ & $54$ & $0$ & Multiclass & 1024 \\
MI & MSLR-WEB10K (Fold 1) & $723412$ & $235259$ & $241521$ & $136$ & $0$ & Regression & 1024 \\
\bottomrule
\end{tabular}}
    \label{tab:dataset_info}
\end{table}

We report ROC-AUC for all binary classification datasets, accuracy for multi-class classification datasets and RMSE for regression datasets, with OT being the one exception, where we report log-loss, as it was used as a default metric in the corresponding Kaggle competition. We use the quantile-transform from the Scikit-learn library \cite{scikit-learn} to preprocess the numerical features for all datasets except OT, where the absence of such transformation was shown to be superior \cite{gorishniy2022embeddings}. Additional information about the datasets is provided in \autoref{sec:app_datasest}.

\textbf{Models.} We use MLP as a simple deep baseline to compare and ablate the methods. Our implementation of MLP exactly follows \cite{gorishniy2021revisiting}, the model is regularized by dropout and weight decay. As more advanced deep models, we evaluate MLP equipped with numerical feature embeddings, specifically, target-aware piecewise linear encoding (MLP-T-LR) and embeddings with periodic activations (MLP-PLR) from \cite{gorishniy2022embeddings}. These models represent the current state-of-the-art solution for tabular DL \cite{gorishniy2022embeddings}, and are of interest as most prior work on pretraining in tabular DL focus on pretraining with the simplest MLP models in evaluation. The implementation of models with numerical embeddings follows \cite{gorishniy2022embeddings}. We use AdamW \cite{adamw} optimizer, do not use learning rate schedules and fix batch sizes for each dataset based on the dataset size.

\textbf{Pretraining.} Pretraining is always performed directly on the target dataset and does not exploit additional data. The learning process thus comprises two stages. On the first stage, the model parameters are optimized w.r.t. the pretraining objective. On the second stage, the model is initialized with the pretrained weights and finetuned on the downstream classification or regression task. We focus on the fully-supervised setup, i.e., assume that target labels are provided for all dataset objects. Typically, pretraining stage involves the input corruption: for instance, to generate positive pairs in contrastive-like objectives or to corrupt the input for reconstruction in self-prediction based objectives. We use random feature resampling as a proven simple baseline for input corruption in tabular data \cite{scarf, vime}.
Learning rate and weight decay are shared between the two stages (see \autoref{tab:split_share_lr_wd} for the ablation). We fix the maximum number of pretraining iterations for each dataset at $100k$. On every $10k$-th iteration, we compute the value of the pretraining objective using the hold-out validation objects for early-stopping on large-scale WE, CO and MI datasets. On other datasets we directly finetune the current model every $10k$-th iteration and perform early-stopping based on the target metric after finetuning (we do not observe much difference between early stopping by loss or by downstream metric, see \autoref{tab:early_stop}).

\textbf{Hyperparameters \& Evaluation.} Hyperparameter tuning is crucial for a fair comparison, therefore, we use Optuna \cite{optuna} to optimize the model and pretraining hyperparameters for each method on each dataset. We use the validation subset of each dataset for hyperparameter tuning. The exact search spaces for the hyperparameters of each method are provided in \autoref{sec:app_hyperparameters}.

We run the tuned configuration of each pretraining method with $15$ random seeds and report the average metric on the test splits. When comparing to GBDT, we  obtain three ensembles by splitting the fifteen single model predictions into three disjoint subsets of five models and averaging predictions within each subset. Then, we report the average metric over the three ensembles.

\subsection{Comparing pretraining objectives}

Here we compare the contrastive learning and self-prediction objectives from prior work in the described setup. For contrastive learning, we follow the method described in \cite{scarf}: use InfoNCE loss, consider corrupted inputs $\hat{x}$ as positives for $x$ and the rest of the batch as negatives.
For self-prediction methods, we evaluate two objectives: the first one is the reconstruction of the original $x$, given the corrupted input $\hat{x}$ (the reconstruction loss is computed for all columns), the second one is the binary mask prediction, where the objective is to predict the mask vector $m$ indicating the corrupted columns from the corrupted input $\hat{x}$. The results of the comparison are in \autoref{tab:compare_objectives}. We summarize our key findings below.

\textbf{Contrastive is not superior.} Both the reconstruction and the mask prediction objectives are preferable to the contrastive objective.
The two self-prediction objectives have the advantage of being conceptually simpler, and easier to implement, while also being less resource-intensive (no need for the second view of augmented examples in each batch, simpler loss function).
We thus recommend the self-prediction based objectives as a practical solution for pretraining in tabular DL.

\textbf{Pretraining is beneficial for the state-of-the-art models.} Models with the numerical feature embeddings also benefit from pretraining with either reconstruction or mask prediction demonstrating the top performance on the downstream task. However, the improvement is typically less noticeable compared to the vanilla MLPs.

\textbf{There is no universal solution between self-prediction objectives.} We observe that for some datasets the reconstruction objective outperforms the mask prediction (OT, WE, CO, MI), while on others the mask prediction is better (GE, CH, HI, AD). We also note that the mask prediction objective sometimes leads to unexpected performance drops for models with numerical embeddings (WE, MI), we do not observe significant performance drops for the reconstruction objective.

\textbf{The main takeaway:} simple pretraining strategies based on self-prediction lead to significant improvements in the downstream accuracy compared to the tuned supervised baselines learned from scratch across different tabular DL models and datasets.
In practice, we recommend trying both reconstruction and mask prediction as tabular pretraining baselines, as either one might show superior performance depending on the dataset being used.

\begin{table}[h!]
    \centering
    \caption{Results for pretraining deep models with different objectives. We report metrics averaged over 15 seeds, bold entries correspond to results that are statistically significantly better (we use Tukey HSD test). The comparisons are separate for different models. \textuparrow\ corresponds to accuracy and ROC-AUC metrics, \textdownarrow\ corresponds to RMSE and log-loss for OT. "no pretraining" stands for the supervised baseline, initialized with random weights}
    \begin{tabular}{lccccccccccc}
\toprule
{} & GE \textuparrow & CH \textuparrow & CA \textdownarrow & HO \textdownarrow & OT \textdownarrow & HI \textuparrow & FB \textdownarrow & AD \textuparrow & WE \textdownarrow & CO \textuparrow & MI \textdownarrow \\
\midrule
\multicolumn{12}{c}{MLP} \\
\midrule
no pretraining         & $0.635$ & $0.849$ & $0.506$ & $3.156$ & $0.479$ & $0.801$ & $5.737$ & $0.908$ & $1.909$ & $0.963$ & $0.749$ \\
contrastive     & $0.672$ & $\mathbf{0.855}$ & $0.455$ & $\mathbf{3.056}$ & $0.469$ & $\mathbf{0.813}$ & $5.697$ & $0.910$ & $1.881$ & $0.960$ & $0.748$ \\
rec             & $0.662$ & $0.853$ & $\mathbf{0.445}$ & $\mathbf{3.044}$ & $\mathbf{0.466}$ & $0.805$ & $\mathbf{5.641}$ & $\mathbf{0.910}$ & $\mathbf{1.875}$ & $\mathbf{0.965}$ & $\mathbf{0.746}$ \\
mask            & $\mathbf{0.691}$ & $\mathbf{0.857}$ & $0.454$ & $3.113$ & $0.472$ & $\mathbf{0.814}$ & $\mathbf{5.681}$ & $\mathbf{0.912}$ & $1.883$ & $0.964$ & $0.748$ \\
\midrule
\multicolumn{12}{c}{MLP-PLR} \\
\midrule
no pretraining         & $0.668$ & $\mathbf{0.858}$ & $0.469$ & $\mathbf{3.008}$ & $0.483$ & $0.809$ & $\mathbf{5.608}$ & $0.926$ & $1.890$ & $0.969$ & $0.746$ \\
rec             & $0.667$ & $0.852$ & $\mathbf{0.439}$ & $\mathbf{3.031}$ & $\mathbf{0.472}$ & $0.808$ & $\mathbf{5.571}$ & $\mathbf{0.926}$ & $\mathbf{1.877}$ & $\mathbf{0.971}$ & $\mathbf{0.745}$ \\
mask            & $\mathbf{0.685}$ & $\mathbf{0.863}$ & $\mathbf{0.434}$ & $\mathbf{3.007}$ & $0.477$ & $\mathbf{0.818}$ & $\mathbf{5.586}$ & $\mathbf{0.927}$ & $1.911$ & $\mathbf{0.970}$ & $0.748$ \\
\midrule
\multicolumn{12}{c}{MLP-T-LR} \\
\midrule
no pretraining         & $0.634$ & $\mathbf{0.866}$ & $0.444$ & $3.113$ & $0.482$ & $0.805$ & $5.520$ & $0.925$ & $1.897$ & $0.968$ & $0.749$ \\
rec             & $\mathbf{0.652}$ & $0.857$ & $\mathbf{0.424}$ & $3.109$ & $\mathbf{0.472}$ & $0.808$ & $\mathbf{5.363}$ & $0.924$ & $\mathbf{1.861}$ & $\mathbf{0.969}$ & $\mathbf{0.746}$ \\
mask            & $\mathbf{0.654}$ & $\mathbf{0.868}$ & $\mathbf{0.424}$ & $\mathbf{3.045}$ & $\mathbf{0.472}$ & $\mathbf{0.818}$ & $5.544$ & $\mathbf{0.926}$ & $1.916$ & $\mathbf{0.969}$ & $0.748$ \\
\bottomrule
\end{tabular}

    \label{tab:compare_objectives}
\end{table}

\section{Target-aware pretraining objectives}

In this section, we show that exploiting the target variables during the pretraining stage can further increase the downstream performance. Specifically, we evaluate several strategies to leverage information about targets during pretraining, identify the best ones and compare them to GBDT. Below we describe a list of target-aware pretraining objectives that we investigate.

\begin{table}[h!]
    \centering
    \caption{Variations of the target-aware pretraining schemes. Notation follows \autoref{tab:compare_objectives}. Bold results indicate statistically significant winners across all models and methods. "+ target" denotes target-conditioned pretraining, "+ sup" denotes auxiliary supervised head.}
    {\small \begin{adjustbox}{center}
\begin{tabular}{lccccccccccc|c}
\toprule
{} & GE \textuparrow & CH \textuparrow & CA \textdownarrow & HO \textdownarrow & OT \textdownarrow & HI \textuparrow & FB \textdownarrow & AD \textuparrow & WE \textdownarrow & CO \textuparrow & MI \textdownarrow & Avg. Rank \\
\midrule
\multicolumn{12}{c}{MLP} \\
\midrule
no pretraining         & $0.635$ & $0.849$ & $0.506$ & $3.156$ & $0.479$ & $0.801$ & $5.737$ & $0.908$ & $1.909$ & $0.963$ & $0.749$ & $5.5 \pm 1.4$
\\
mask            & $0.691$ & $0.857$ & $0.454$ & $3.113$ & $0.472$ & $0.814$ & $5.681$ & $0.912$ & $1.883$ & $0.964$ & $0.748$ & $3.8 \pm 1.4$
\\
rec             & $0.662$ & $0.853$ & $0.445$ & $\mathbf{3.044}$ & $0.466$ & $0.805$ & $5.641$ & $0.910$ & $1.875$ & $0.965$ & $0.746$ & $3.6 \pm 1.5$
\\
sup             & $0.693$ & $0.856$ & $0.441$ & $3.077$ & $\mathbf{0.459}$ & $0.814$ & $5.689$ & $0.914$ & $1.883$ & $0.968$ & $0.748$ & $3.0 \pm 1.0$
\\
mask + target   & $0.683$ & $0.857$ & $0.434$ & $\mathbf{3.056}$ & $0.468$ & $\mathbf{0.819}$ & $5.633$ & $0.914$ & $1.876$ & $0.965$ & $0.748$ & $2.9 \pm 1.3$
\\
rec + target    & $0.659$ & $0.853$ & $0.454$ & $\mathbf{3.044}$ & $0.463$ & $0.806$ & $5.636$ & $0.909$ & $1.884$ & $0.965$ & $\mathbf{0.745}$ & $3.7 \pm 1.9$
\\
mask + sup        & $0.693$ & $0.857$ & $0.436$ & $3.099$ & $\mathbf{0.458}$ & $0.817$ & $5.685$ & $0.915$ & $1.873$ & $0.967$ & $0.748$ & $2.7 \pm 1.2$
\\
rec + sup         & $0.684$ & $0.854$ & $0.436$ & $\mathbf{3.012}$ & $\mathbf{0.456}$ & $0.815$ & $5.672$ & $0.911$ & $\mathbf{1.862}$ & $0.967$ & $0.747$ & $2.6 \pm 1.5$
\\
\midrule
\multicolumn{12}{c}{MLP-PLR} \\
\midrule
no pretraining         & $0.668$ & $0.858$ & $0.469$ & $\mathbf{3.008}$ & $0.483$ & $0.809$ & $5.608$ & $0.926$ & $1.890$ & $0.969$ & $0.746$ & $3.5 \pm 1.7$
\\
mask            & $0.685$ & $\mathbf{0.863}$ & $0.434$ & $\mathbf{3.007}$ & $0.477$ & $0.818$ & $5.586$ & $0.927$ & $1.911$ & $0.970$ & $0.748$ & $2.8 \pm 1.7$
\\
rec             & $0.667$ & $0.852$ & $0.439$ & $\mathbf{3.031}$ & $0.472$ & $0.808$ & $5.571$ & $0.926$ & $1.877$ & $\mathbf{0.971}$ & $\mathbf{0.745}$ & $2.6 \pm 1.2$
\\
sup             & $\mathbf{0.710}$ & $0.859$ & $0.433$ & $3.136$ & $0.479$ & $0.811$ & $5.521$ & $0.924$ & $1.873$ & $\mathbf{0.971}$ & $0.748$ & $2.5 \pm 1.2$
\\
mask + target   & $0.694$ & $0.862$ & $\mathbf{0.425}$ & $\mathbf{3.023}$ & $0.474$ & $\mathbf{0.821}$ & $5.537$ & $\mathbf{0.929}$ & $1.911$ & $0.969$ & $0.749$ & $2.5 \pm 1.9$
\\
rec + target    & $0.688$ & $0.860$ & $0.445$ & $\mathbf{3.064}$ & $0.475$ & $0.812$ & $5.507$ & $0.927$ & $1.887$ & $\mathbf{0.971}$ & $0.748$ & $2.7 \pm 1.3$
\\
mask + sup       & $\mathbf{0.711}$ & $\mathbf{0.866}$ & $0.441$ & $3.129$ & $0.480$ & $0.813$ & $5.480$ & $0.925$ & $1.875$ & $0.969$ & $\mathbf{0.745}$ & $2.5 \pm 1.4$
\\
rec + sup         & $\mathbf{0.709}$ & $0.858$ & $0.433$ & $\mathbf{3.059}$ & $0.465$ & $0.807$ & $5.571$ & $0.927$ & $\mathbf{1.865}$ & $\mathbf{0.971}$ & $\mathbf{0.745}$ & $1.9 \pm 1.2$
\\
\midrule
\multicolumn{12}{c}{MLP-T-LR} \\
\midrule
no pretraining         & $0.634$ & $\mathbf{0.866}$ & $0.444$ & $3.113$ & $0.482$ & $0.805$ & $5.520$ & $0.925$ & $1.897$ & $0.968$ & $0.749$ & $3.9 \pm 1.7$
\\
mask            & $0.654$ & $\mathbf{0.868}$ & $\mathbf{0.424}$ & $\mathbf{3.045}$ & $0.472$ & $0.818$ & $5.544$ & $0.926$ & $1.916$ & $0.969$ & $0.748$ & $2.8 \pm 1.7$
\\
rec             & $0.652$ & $0.857$ & $\mathbf{0.424}$ & $3.109$ & $0.472$ & $0.808$ & $\mathbf{5.363}$ & $0.924$ & $\mathbf{1.861}$ & $0.969$ & $\mathbf{0.746}$ & $2.5 \pm 1.4$
\\
sup             & $0.682$ & $0.860$ & $0.430$ & $3.135$ & $0.471$ & $0.807$ & $5.525$ & $0.927$ & $1.893$ & $\mathbf{0.971}$ & $0.747$ & $2.8 \pm 1.5$
\\
mask + target   & $0.649$ & $\mathbf{0.865}$ & $\mathbf{0.421}$ & $\mathbf{3.058}$ & $0.474$ & $\mathbf{0.820}$ & $5.644$ & $\mathbf{0.929}$ & $1.924$ & $0.969$ & $0.749$ & $2.8 \pm 2.1$
\\
rec + target    & $0.668$ & $\mathbf{0.864}$ & $0.440$ & $3.113$ & $0.473$ & $0.806$ & $5.493$ & $0.927$ & $\mathbf{1.862}$ & $0.969$ & $\mathbf{0.746}$ & $2.5 \pm 1.4$
\\
mask + sup        & $0.676$ & $0.858$ & $0.429$ & $3.199$ & $0.468$ & $0.814$ & $5.510$ & $0.926$ & $1.869$ & $\mathbf{0.971}$ & $0.748$ & $2.5 \pm 0.8$
\\
rec + sup         & $0.678$ & $\mathbf{0.865}$ & $0.437$ & $3.112$ & $0.462$ & $0.807$ & $5.516$ & $0.927$ & $\mathbf{1.862}$ & $\mathbf{0.970}$ & $0.748$ & $2.4 \pm 1.2$
\\
\bottomrule
\end{tabular}
\end{adjustbox}
}
    \label{tab:target_mlp_variations}
\end{table}

\textbf{Supervised loss with augmentations}. A straightforward way to incorporate the target variable into the pretraining is by using the input corruption as an augmentation for the standard supervised learning objective. An important difference of this baseline in our setup to the one in \cite{scarf} is that we treat learning on corrupted samples as a pretraining stage and finetune the entire model on the full uncorrupted dataset afterwards (we ablate this in \autoref{sec:two_stage}).

\textbf{Supervised loss with augmentations + self-prediction}. We evaluate a natural extension to the above baseline: a combination of the supervised objective with the unsupervised self-prediction.
Note, that during the pretraining stage both losses are calculated on corrupted inputs, while the finetuning is performed on the non-corrupted dataset. For the self-prediction objectives we evaluate both the reconstruction and the mask prediction. We use different prediction heads for supervised and self-prediction objectives. We sum supervised and self-prediction losses with equal weights.

\textbf{Target-aware pretraining}. An alternative to the approaches described above is the modification of the pretraining task itself. An example of this approach is supervised contrastive learning \cite{supcon}, where the target variable is used to sample positive and negative examples. We introduce the target variable into the self-prediction based objectives with two modifications.

First, we condition the mask prediction or the reconstruction head on the original input's target by concatenating the hidden representation from the backbone network $z = f(\hat{x})$ with the target variable representation before passing it to the pretraining head to obtain predictions $p = h(\mathtt{concat}[z, y])$. For classification datasets we encode $y$ with one-hot-encoding, for regression targets we use the standard scaling.

Second, we change the input corruption scheme by sampling the replacement from the feature target conditional distribution where a target is different to the original. Intuitively, corrupting the input object $x$ in the direction of the target different to the original makes the pretraining task more correlated with the downstream target prediction. Concretely, given an object-target pair $(x_{i}, y_{i})$, we sample a new target $\hat{y}_{i}$ from a uniform distribution over the set $\{ y\ |\ y \neq y_{i} \}$ \footnote{\label{note:reg_target}For regression problems, when the target variable is continuous, we preliminarily discretize it into $n$ uniform bins, where $n$ is chosen according to the Freedman–Diaconis rule \cite{freedman-diaconis}.}, then each feature $x^{j}$ is replaced with a sample from the $p(x^{j} | \hat{y}_{i})$ distribution, instead of $p(x^{j})$.

\subsection{Comparing target-aware objectives}

Here we compare the strategies of incorporating the target variable into pretraining.
The results of the comparison are in \autoref{tab:target_mlp_variations}. Our key findings are formulated below.

\textbf{Supervised loss with augmentations is another strong baseline for MLP}. Pretraining with the supervised loss on corrupted data consistently improves over supervised training from scratch for the MLP. This objective is a strong baseline along with the self-prediction based objectives. However, for models with numerical embeddings the supervised objective with corruptions is less consistent and sometimes is inferior to training from scratch, thus for these models we recommend the self-prediction objectives alone as baselines.

\textbf{Target-aware objectives demonstrate the best performance}.
Both the supervised loss with self-prediction and modified self-prediction objectives improve over the unsupervised pretraining baselines across datasets and model architectures.

For the objective with the combination of supervised and self-prediction losses the variation with the reconstruction loss is the most consistent across models and dataset with no performance drops below the pretraining-free baseline. The variant with mask prediction shows similar performance and stability for the MLP, but is not as good as reconstruction for models with numerical embeddings.

For the target-aware self-prediction objectives, the modified mask prediction delivers significant improvements over its unsupervised counterpart. Modified reconstruction objective, however, does not improve over unsupervised reconstruction objective.

\textbf{Main takeaways}:
Target-aware objectives help further increase the downstream performance, improving upon their unsupervised counterparts. For the reconstruction based self-prediction baseline the addition of the supervised loss is most beneficial ("rec + sup" from \autoref{tab:target_mlp_variations}), for the mask prediction objective it's target-aware modification provides more improvements ("mask + target" from \autoref{tab:target_mlp_variations}).
A simple MLP model pretrained with those "target-aware" objectives often reaches or surpasses complex models with numerical embeddings trained from scratch.
In practice, we recommend first trying the baseline pretraining objectives ("rec", "mask", "sup" from \autoref{tab:target_mlp_variations}), choosing the suitable baseline for the dataset and improving it accordingly: supervised loss for the reconstruction and target-aware modification for the mask prediction.

\subsection{Comparison to GBDT}

Here we compare MLPs and MLPs with numerical feature embeddings pretrained with the supervised loss with reconstruction and the target-aware mask prediction objectives to the GBDTs. \autoref{tab:the_comparison} shows the results of the comparison.

\begin{table}[h!]
    \centering
    \caption{Comparison of pretrained models to GBDT. Notation follows \autoref{tab:compare_objectives}. Results represent ensembles of models. Bold entries correspond to the overall statistically significant best entries.}
    {\small \begin{adjustbox}{center}
\begin{tabular}{lccccccccccc|c}
\toprule
{} & GE \textuparrow & CH \textuparrow & CA \textdownarrow & HO \textdownarrow & OT \textdownarrow & HI \textuparrow & FB \textdownarrow & AD \textuparrow & WE \textdownarrow & CO \textuparrow & MI \textdownarrow & Avg. Rank
\\
\midrule
CatBoost         & $0.692$ & $\mathbf{0.864}$ & $0.430$ & $3.093$ & $0.450$ & $0.807$ & $5.226$ & $0.928$ & $1.801$ & $0.967$ & $\mathbf{0.741}$ & $2.6 \pm 1.4$
\\
XGBoost         & $0.683$ & $0.860$ & $0.434$ & $3.152$ & $0.454$ & $0.805$ & $5.338$ & $0.927$ & $\mathbf{1.782}$ & $0.969$ & $\mathbf{0.742}$ & $3.0 \pm 1.5$
\\
\midrule
\multicolumn{12}{c}{MLP} \\
\midrule
no pretraining         & $0.656$ & $0.852$ & $0.482$ & $3.055$ & $0.467$ & $0.805$ & $5.666$ & $0.910$ & $1.850$ & $0.968$ & $0.747$ & $4.8 \pm 1.1$
\\
mask + target   & $0.709$ & $0.860$ & $0.414$ & $2.949$ & $0.457$ & $\mathbf{0.828}$ & $5.551$ & $0.916$ & $1.809$ & $0.969$ & $0.746$ & $2.8 \pm 1.2$
\\
rec + sup         & $0.709$ & $0.859$ & $0.419$ & $2.951$ & $\mathbf{0.442}$ & $0.817$ & $5.531$ & $0.913$ & $1.801$ & $0.973$ & $0.745$ & $2.5 \pm 1.2$
\\
\midrule
\multicolumn{12}{c}{MLP-P-LR} \\
\midrule
no pretraining         & $0.695$ & $\mathbf{0.864}$ & $0.454$ & $2.953$ & $0.470$ & $0.814$ & $5.324$ & $0.928$ & $1.835$ & $\mathbf{0.974}$ & $0.744$ & $2.6 \pm 1.2$
\\
mask + target   & $\mathbf{0.719}$ & $\mathbf{0.866}$ & $\mathbf{0.407}$ & $2.952$ & $0.458$ & $\mathbf{0.828}$ & $5.373$ & $\mathbf{0.930}$ & $1.849$ & $0.973$ & $0.745$ & $2.1 \pm 1.2$
\\
rec + sup         & $\mathbf{0.737}$ & $0.862$ & $0.424$ & $2.964$ & $0.449$ & $0.811$ & $\mathbf{5.124}$ & $\mathbf{0.929}$ & $1.813$ & $\mathbf{0.974}$ & $0.744$ & $2.0 \pm 1.0$
\\
\midrule
\multicolumn{12}{c}{MLP-T-LR} \\
\midrule
no pretraining         & $0.662$ & $\mathbf{0.868}$ & $0.437$ & $3.028$ & $0.472$ & $0.808$ & $5.424$ & $0.927$ & $1.850$ & $0.972$ & $0.747$ & $3.7 \pm 1.1$
\\
mask + target   & $0.673$ & $\mathbf{0.868}$ & $\mathbf{0.410}$ & $\mathbf{2.894}$ & $0.460$ & $\mathbf{0.827}$ & $5.458$ & $\mathbf{0.930}$ & $1.849$ & $0.972$ & $0.746$ & $2.4 \pm 1.4$
\\
rec + sup         & $0.705$ & $\mathbf{0.866}$ & $0.425$ & $3.057$ & $\mathbf{0.444}$ & $0.814$ & $5.422$ & $0.927$ & $1.811$ & $\mathbf{0.974}$ & $0.746$ & $2.5 \pm 1.1$
\\
\bottomrule
\end{tabular}
\end{adjustbox}}
    \label{tab:the_comparison}
\end{table}

We observe that both pretraining with target aware objectives and using numerical feature embeddings consistently improve the performance of the simple MLP backbone. In particular, MLP coupled with target aware pretraining starts to outperform GBDT on 4 datasets (GE, CA, OT, HI). Combined with numerical feature embeddings, pretraining improves MLP performance further, making it superior to GBDT on the majority of the datasets, with two exceptions in WE and MI.

\section{Analysis}

\subsection{Investigating the properties of pretrained models}

In this section, we provide a possible explanation of why the incorporation of the target variable into pretraining can lead to better downstream task performance. We do this through the experiments on the controllable synthetic data. Here we describe the properties and the generation process of the data and our observations on the differences of the pretraining schemes.

\begin{figure}[h!]
    \centering
    \includegraphics[width=0.6\textwidth]{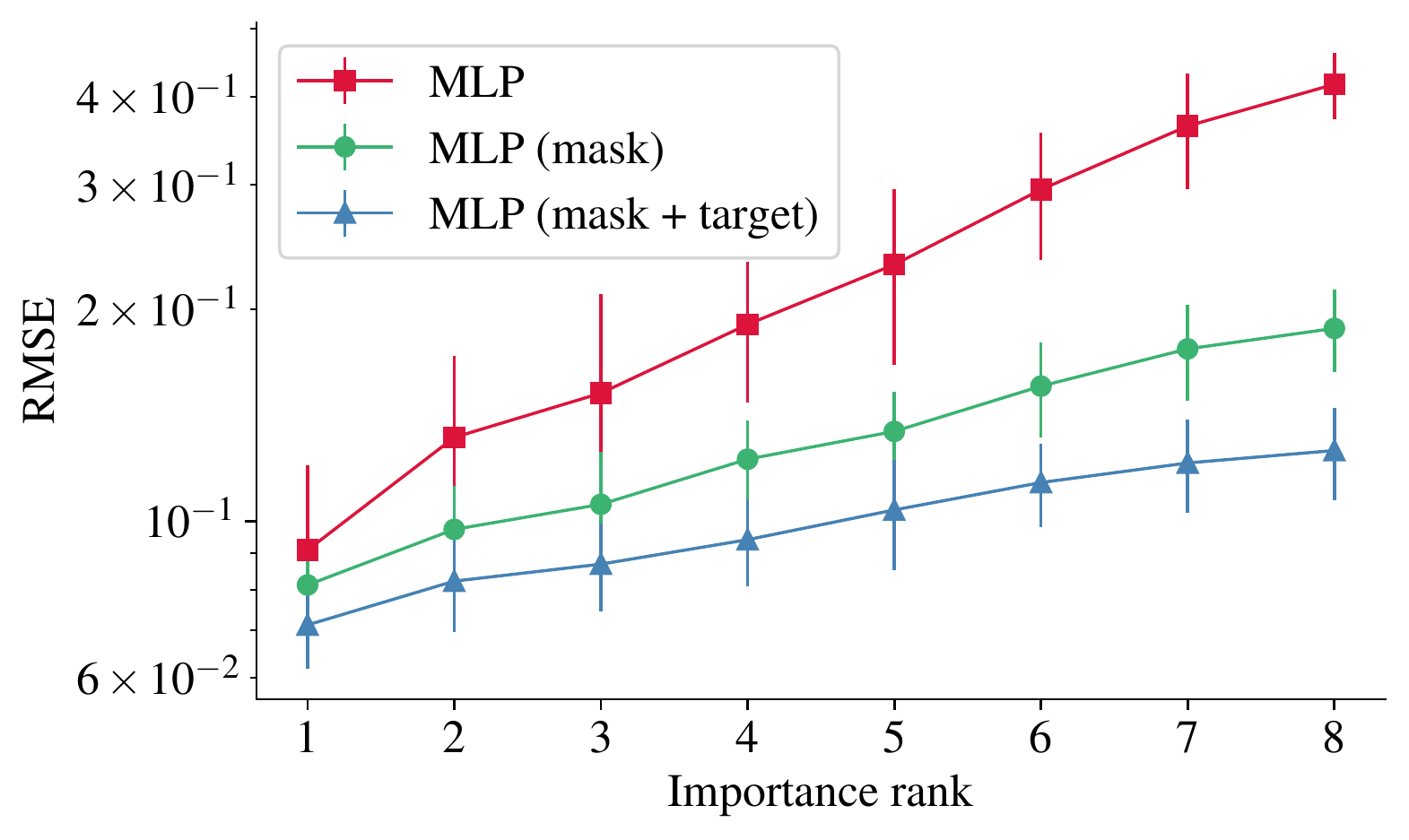}
    \caption{The decodability of object feature from the intermediate representations computed by the pretrained models and the models trained from scratch. The pretrained models decently capture the information about all the features, while the randomly initialized models capture the most informative features and suppress the others.}
    \label{fig:target_analysis}
\end{figure}

We follow the synthetics generation protocol described in \cite{gorishniy2021revisiting} with a modification that allows for the manual control of the feature importance for the particular prediction task. Concretely, we generate the objects features $\lrf{x_{i}}_{i=1}^{n}$ as samples from the multivariate Gaussian distribution with zero mean and covariance matrix $\Sigma$ with identical diagonal and a constant $c = 0.5$  everywhere else. To generate the corresponding objects targets $\lrf{y_{i}}_{i=1}^n$ we sample a vector $p \in \R^{m}$ from the Dirichlet distribution $p \sim \text{Dir} (1_{m})$ and let $p$ define the influence of the objects features on the target. Then, we build an ensemble of $10$ random oblivious decision trees $\left\{T_i(x)\right\}_{i=1}^{10}$ of depth $10$, where on each tree level we sample a feature $j \sim \text{Cat}(p)$ and a threshold $t \sim \mathrm{Unif}(\mathrm{min}(x^{j}), \mathrm{max}(x^{j}))$ for a decision rule. For each tree leaf, we sample a scalar $l \sim \mathcal{N}(0, 1)$, representing a logit in binary classification. We define the binary targets as follows: $y(x) = \mathrm{I}\left\{\frac{1}{10} \sum_{i=1}^{10} T_{i}(x) > 0 \right\}$.

Intuitively if a particular feature is often used for splitting in the nodes of a decision tree, it would have more influence on the target variable. Indeed, we find the feature importances\footnote{Computed with the CatBoost method ``\texttt{get\_feature\_importance()}''} correlate well with the predefined vector $p$. We set the size of the dataset $n = 50.000$ and the number of features $m = 8$ and generate $50$ datasets with different feature importance vectors $p$ for the analysis.

For each generated dataset, we then check whether the finetuned models capture the information about object features in their intermediate representations. Specifically, we train an MLP to predict the value of  the $i$-th object feature given the frozen embeddings produced by a finetuned network initialized from (a) random initialization, (b) mask prediction pretraining, (c) target-aware mask prediction pretraining. The separate MLP is used for each feature and the RMSE learning objective is used. Then we report the RMSE on the test set for all features $i \in [0, m]$ along with their importance rank in the dataset on \autoref{fig:target_analysis}. Here, the lower ranks correspond to the more important features.

\autoref{fig:target_analysis} reveals that the target-aware pretraining enables the model to capture more information about the informative features compared to the ``unsupervised'' pretraining and, especially, to the learning from scratch. The latter one successfully captures the most informative feature from the training data, while suppressing the less important, but still significant features. We conjecture that this is the source of superiority of the target-aware pretraining.

\subsection{Efficient ensembling}

In this section we show, that it is possible to construct ensembles from one pretraining checkpoint (pretrained with the target conditioned mask prediction objective). To this end, we run finetuning with $15$ different random seeds starting from the one pretrained checkpoint. \autoref{tab:ensembling_finetunes} shows the results.

\begin{table}[h!]
    \centering
    \caption{Efficient ensembling for MLP {\footnotesize mask + target}}
    \begin{tabular}{lccccccccccc}
\toprule
{} & GE \textuparrow & CH \textuparrow & CA \textdownarrow & HO \textdownarrow & OT \textdownarrow & HI \textuparrow & FB \textdownarrow & AD \textuparrow & WE \textdownarrow & CO \textuparrow & MI \textdownarrow \\
\midrule
single & $0.683$ & $0.857$ & $0.434$ & $3.056$ & $0.468$ & $0.819$ & $5.633$ & $0.914$ & $1.876$ & $0.965$ & $0.748$ \\
standard ensemble  & $0.709$ & $0.860$ & $0.414$ & $2.949$ & $0.457$ & $0.828$ & $5.551$ & $0.916$ & $1.809$ & $0.969$ & $0.746$ \\
efficient ensemble & $0.702$ & $0.861$ & $0.411$ & $2.967$ & $0.461$ & $0.825$ & $5.590$ & $0.917$ & $1.820$ & $0.969$ & $0.746$ \\
\bottomrule
\end{tabular}
    \label{tab:ensembling_finetunes}
\end{table}

Both ensembling the models from a shared pretrain checkpoint and ensembling multiple independent pretraining runs produces strong ensembles, which shows that it is sufficient to pretrain once and create ensembles by several independent finetuning processes. This is important in practice since finetuninig is typically cheaper (i.e. requires fewer iterations), and still is able to produce diverse models from the one pretraining checkpoint for ensembles of comparable quality.

\subsection{On importance of finetuning on clean data}
\label{sec:two_stage}

Here we show, that the second stage of finetuning the model on the entire dataset without input corruption is often necessary for the best downstream performance. To this end we compare finetuning the models on clean data with using models right after pretraining for two objectives: supervised loss and supervised loss with the reconstruction objective.

\begin{table}[h!]
    \centering
    \caption{Finetuning MLP on clean data versus using the model trained on corrupted inputs only}
    \begin{tabular}{lccccccccccc}
\toprule
{} & GE \textuparrow & CH \textuparrow & CA \textdownarrow & HO \textdownarrow & OT \textdownarrow & HI \textuparrow & FB \textdownarrow & AD \textuparrow & WE \textdownarrow & CO \textuparrow & MI \textdownarrow \\
\midrule
no pretraining         & $0.635$ & $0.849$ & $0.506$ & $3.156$ & $0.479$ & $0.801$ & $5.737$ & $0.908$ & $1.909$ & $0.963$ & $0.749$ \\
\midrule
sup             & $0.693$ & $0.856$ & $0.441$ & $3.077$ & $0.459$ & $0.814$ & $5.689$ & $0.914$ & $1.883$ & $0.968$ & $0.748$ \\
sup | no finetune            & $0.674$ & $0.853$ & $0.464$ & $3.251$ & $0.461$ & $0.808$ & $6.167$ & $0.910$ & $1.938$ & $0.958$ & $0.752$ \\
\midrule
rec + sup         & $0.684$ & $0.854$ & $0.436$ & $3.012$ & $0.456$ & $0.815$ & $5.672$ & $0.911$ & $1.862$ & $0.967$ & $0.747$ \\
rec + sup | no finetune             & $0.683$ & $0.853$ & $0.467$ & $3.232$ & $0.474$ & $0.811$ & $6.044$ & $0.907$ & $1.901$ & $0.956$ & $0.752$ \\
\bottomrule
\end{tabular}

    \label{tab:finetune_important}
\end{table}

Across all datasets for both methods finetuning on uncorrupted data with the supervised loss proves to be essential for the best performance. Sometimes excluding the second finetuning stage degrades the performance below the tuned supervised baseline of training from scratch.

\subsection{Does pretraining require more compute?}

In this section we investigate how much more time is spent on pretraining, compared to training the models from scratch. We run pretraining with "rec + sup" objective with 50k, 100k and 150k pretraining iterations thresholds (early-stopping, in theory, could make 100k and more iterations equivalent to the 50k, but in practice it was not the case). We report downstream performance along with average time spent to pretrain and finetune a model in \autoref{tab:pretraining_time}.

\begin{table}[h!]
    \centering
    \caption{Comparison of time spent for MLP training from scratch and "rec + sup" pretraining with 50k, 100k and 150k max-iterations threshold. Second row in each group reports average time spent training one model in seconds on an A100 GPU.}
    \begin{tabular}{lccccccccccc}
\toprule
{} & GE \textuparrow & CH \textuparrow & CA \textdownarrow & HO \textdownarrow & OT \textdownarrow & HI \textuparrow & FB \textdownarrow & AD \textuparrow & WE \textdownarrow & CO \textuparrow & MI \textdownarrow \\
\midrule
no pretraining         & $0.635$ & $0.849$ & $0.506$ & $3.156$ & $0.479$ & $0.801$ & $5.737$ & $0.908$ & $1.909$ & $0.963$ & $0.749$ \\
                & {\footnotesize 29s} & {\footnotesize 10s} & {\footnotesize 25s} & {\footnotesize 26s} & {\footnotesize 38s} & {\footnotesize 29s} & {\footnotesize 159s} & {\footnotesize 12s} & {\footnotesize 57s} & {\footnotesize 352s} & {\footnotesize 211s} 
\\
\midrule
rec + sup | 50k & $0.679$ & $0.857$ & $0.441$ & $3.064$ & $0.462$ & $0.813$ & $5.650$ & $0.910$ & $1.879$ & $0.966$ & $0.747$ \\
                & {\footnotesize 327s} & {\footnotesize 227s} & {\footnotesize 280s} & {\footnotesize 277s} & {\footnotesize 355s} & {\footnotesize 256s} & {\footnotesize 624s} & {\footnotesize 403s} & {\footnotesize 242s} & {\footnotesize 593s} & {\footnotesize 292s}
\\
\midrule
rec + sup | 100k         & $0.684$ & $0.854$ & $0.436$ & $3.012$ & $0.456$ & $0.815$ & $5.672$ & $0.911$ & $1.862$ & $0.967$ & $0.747$ \\
                & {\footnotesize 661s} & {\footnotesize 312s} & {\footnotesize 533s} & {\footnotesize 455s} & {\footnotesize 570s} & {\footnotesize 526s} & {\footnotesize 740s} & {\footnotesize 759s} & {\footnotesize 439s} & {\footnotesize 649s} & {\footnotesize 472s} 
\\
\midrule
rec + sup | 150k & $0.692$ & $0.859$ & $0.435$ & $3.012$ & $0.456$ & $0.816$ & $5.629$ & $0.910$ & $1.866$ & $0.968$ & $0.746$ \\
                & {\footnotesize 891s} & {\footnotesize 338s} & {\footnotesize 758s} & {\footnotesize 509s} & {\footnotesize 712s} & {\footnotesize 862s} & {\footnotesize 916s} & {\footnotesize 1069s} & {\footnotesize 663s} & {\footnotesize 927s} & {\footnotesize 665s} 
\\
\bottomrule
\end{tabular}
    \label{tab:pretraining_time}
\end{table}

Pretraining often requires by an order of magnitude more compute, it is especially apparent on smaller scale datasets like GE, CH, CA, HO, OT, HI, AD. However, the absolute time spent on pretraining is still acceptable, as the original training from scratch takes seconds on small datasets. Generally the more iterations you use for pretraining, the better downstream quality you get.

\section{Conclusion}

In this work, we have systematically evaluated typical pretraining objectives for tabular deep learning. We have revealed several important recipes for optimal pretraining performance that can be universally beneficial across various problems and models. Our findings confirm that pretraining can significantly improve the performance of tabular deep models and provide additional evidence that tabular DL can become a strong alternative to GBDT.

\bibliographystyle{plain}
\bibliography{references.bib}

\newpage

\appendix

\section{Datasets}
\label{sec:app_datasest}

We used the following datasets:
\begin{itemize}
    \item Gesture Phase Prediction (\cite{gesture})
    \item Churn Modeling\footnote{https://www.kaggle.com/shrutimechlearn/churn-modelling}
    \item California Housing (real estate data, \cite{california})
    \item House 16H\footnote{https://www.openml.org/d/574}
    \item Adult (income estimation, \cite{adult})
    \item Otto Group Product Classification\footnote{https://www.kaggle.com/c/otto-group-product-classification-challenge/data}
    \item Higgs (simulated physical particles, \cite{higgs}; we use the version with 98K samples available at the OpenML repository \cite{openml})
    \item Facebook Comments (\cite{fb-comments})
    \item Covertype (forest characteristics, \cite{covertype})
    \item Microsoft (search queries, \cite{microsoft}). We follow the pointwise approach to learning-to-rank and treat this ranking problem as a regression problem.
    \item Weather (temperature, \cite{weather}). We take 10\% of the dataset for our experiments due to the its large size.
\end{itemize}

\section{Hyperparameters}
\label{sec:app_hyperparameters}

\subsection{CatBoost}

We fix and do not tune the following hyperparameters:
\begin{itemize}
    \itemsep0em
    \item $\texttt{early-stopping-rounds} = 50$
    \item $\texttt{od-pval} = 0.001$
    \item $\texttt{iterations} = 2000$
\end{itemize}

For tuning on the MI and CO datasets, we set the \texttt{task\_type} parameter to ``GPU''. In all other cases (including the evaluation on these two datasets), we set this parameter to ``CPU''.

\begin{table}[h!]
\centering
\caption{CatBoost hyperparameter space}
\label{tab:A-catboost-space}

{\renewcommand{\arraystretch}{1.2}
\begin{tabular}{ll}
    \toprule
    Parameter & Distribution \\
    \midrule
    Max depth & $\mathrm{UniformInt[1,10]}$ \\
    Learning rate & $\mathrm{LogUniform}[0.001, 1]$ \\
    Bagging temperature &  $\mathrm{Uniform}[0, 1]$ \\
    L2 leaf reg  & $\mathrm{LogUniform}[1, 10]$ \\
    Leaf estimation iterations &  $\mathrm{UniformInt}[1, 10]$ \\
    \midrule
    \# Iterations & 100 \\
    \bottomrule
\end{tabular}}
\end{table}

\subsection{XGBoost}

We fix and do not tune the following hyperparameters:
\begin{itemize}
    \itemsep0em
    \item $\texttt{booster} = \text{"gbtree"}$
    \item $\texttt{early-stopping-rounds} = 50$
    \item $\texttt{n-estimators} = 2000$
\end{itemize}

\begin{table}[h!]
\centering
\caption{XGBoost hyperparameter space.}
\label{tab:A-xgboost-space}

{\renewcommand{\arraystretch}{1.2}
\begin{tabular}{ll}
    \toprule
    Parameter & Distribution \\
    \midrule
    Max depth & $\mathrm{UniformInt[3,10]}$ \\
    Min child weight & $\mathrm{LogUniform}[0.0001, 100]$ \\
    Subsample & $\mathrm{Uniform}[0.5, 1]$ \\
    Learning rate & $\mathrm{LogUniform}[0.001, 1]$ \\
    Col sample by tree & $\mathrm{Uniform}[0.5, 1]$ \\
    Gamma & $\{0, \mathrm{LogUniform}[0.001, 100]\}$ \\
    Lambda & $\{0, \mathrm{LogUniform}[0.1, 10]\}$ \\
    \midrule
    \# Iterations & 100 \\
    \bottomrule
\end{tabular}}
\end{table}

\subsection{MLP}
We fix and do not tune the following hyperparameters:
\begin{itemize}
    \itemsep0em
    \item $\texttt{Layer size} = 512$
    \item $\texttt{Head hidden size} = 512$
\end{itemize}

\begin{table}[h!]
\centering
\caption{MLP hyperparameter space. }
\label{tab:A-mlp-space}

{\renewcommand{\arraystretch}{1.2}
\begin{tabular}{ll}
    \toprule
    Parameter & Distribution \\
    \midrule
    \# Layers & $\mathrm{UniformInt}[1,8]$ \\
    Dropout &   $\{0, \mathrm{Uniform}[0, 0.5]\}$ \\
    Learning rate &  $\mathrm{LogUniform}[5e\text{-}5, 0.005]$ \\
    Weight decay &  $\{0, \mathrm{LogUniform}[1e\text{-}6, 1e\text{-}3] \}$\\
    Corrupt Probability  & $\{0, \mathrm{Uniform}[0.2, 0.8]\}$ \\
    \midrule
    \# Iterations & 100 \\
    \bottomrule
\end{tabular}}
\end{table}

\subsection{Embedding Hyperparameters}
We fix and do not tune the following hyperparameters:
\begin{itemize}
    \itemsep0em
    \item $\texttt{Layer size} = 512$
    \item $\texttt{Head hidden size} = 512$
\end{itemize}

The distribution for the output dimensions of linear layers is $\mathrm{UniformInt}[1, 128]$.

\texttt{PLR, T-LR}. We share the same hyperparameter space for models with embeddings across all datasets.

For the target-aware embeddings (tree-based) \texttt{T-LR}, the distribution for the number of leaves is $\mathrm{UniformInt}[2,256]$, the distribution for the minimum number of items per leaf is $\mathrm{UniformInt}[1,128]$ and the distribution for the minimum information gain required for making a split is $\mathrm{LogUniform}[1e\text{-}9, 0.01]$.

For the periodic embeddings \texttt{PLR}. The distribution for $k$ is $\mathrm{UniformInt}[1,128]$, the distribution for the $\sigma$ parameter is $\mathrm{LogUniform}[0.01, 100]$

\section{Share or split learning rate and weight decay between pretraining and finetuning?}

Here we demonstrate that tuning and using the same learning rate and weight decay for both pretraining and finetuning results in similar performance to tuning these parameters separately for the two stages. We opt for sharing the learning rate and weight decay for pretraining and finetuning in all the experiments in the paper.

\begin{table}[h!]
    \centering
    \caption{Results for single models with MLP {\footnotesize mask + target} pretraining}
    {\small \begin{adjustbox}{center}
\begin{tabular}{lccccccccccc}
\toprule
{} & GE \textuparrow & CH \textuparrow & CA \textdownarrow & HO \textdownarrow & OT \textdownarrow & HI \textuparrow & FB \textdownarrow & AD \textuparrow & WE \textdownarrow & CO \textuparrow & MI \textdownarrow \\
\midrule
shared wd / shared lr  & $0.683 \scriptscriptstyle \pm \scriptstyle 1e\text{-}2$ & $0.857 \scriptscriptstyle \pm \scriptstyle 2e\text{-}3$ & $0.434 \scriptscriptstyle \pm \scriptstyle 7e\text{-}3$ & $3.056 \scriptscriptstyle \pm \scriptstyle 4e\text{-}2$ & $0.468 \scriptscriptstyle \pm \scriptstyle 2e\text{-}3$ & $0.819 \scriptscriptstyle \pm \scriptstyle 2e\text{-}3$ & $5.633 \scriptscriptstyle \pm \scriptstyle 4e\text{-}2$ & $0.914 \scriptscriptstyle \pm \scriptstyle 1e\text{-}3$ & $1.876 \scriptscriptstyle \pm \scriptstyle 5e\text{-}3$ & $0.965 \scriptscriptstyle \pm \scriptstyle 7e\text{-}4$ & $0.748 \scriptscriptstyle \pm \scriptstyle 4e\text{-}4$ \\
shared wd / split lr & $0.697 \scriptscriptstyle \pm \scriptstyle 9e\text{-}3$ & $0.857 \scriptscriptstyle \pm \scriptstyle 3e\text{-}3$ & $0.431 \scriptscriptstyle \pm \scriptstyle 7e\text{-}3$ & $3.032 \scriptscriptstyle \pm \scriptstyle 3e\text{-}2$ & $0.469 \scriptscriptstyle \pm \scriptstyle 2e\text{-}3$ & $0.819 \scriptscriptstyle \pm \scriptstyle 2e\text{-}3$ & $5.647 \scriptscriptstyle \pm \scriptstyle 4e\text{-}2$ & $0.915 \scriptscriptstyle \pm \scriptstyle 9e\text{-}4$ & $1.934 \scriptscriptstyle \pm \scriptstyle 8e\text{-}3$ & $0.964 \scriptscriptstyle \pm \scriptstyle 9e\text{-}4$ & $0.748 \scriptscriptstyle \pm \scriptstyle 4e\text{-}4$ \\
split wd / split lr & $0.688 \scriptscriptstyle \pm \scriptstyle 9e\text{-}3$ & $0.856 \scriptscriptstyle \pm \scriptstyle 3e\text{-}3$ & $0.430 \scriptscriptstyle \pm \scriptstyle 4e\text{-}3$ & $3.046 \scriptscriptstyle \pm \scriptstyle 4e\text{-}2$ & $0.471 \scriptscriptstyle \pm \scriptstyle 3e\text{-}3$ & $0.821 \scriptscriptstyle \pm \scriptstyle 7e\text{-}4$ & $5.734 \scriptscriptstyle \pm \scriptstyle 5e\text{-}2$ & $0.914 \scriptscriptstyle \pm \scriptstyle 7e\text{-}4$ & $1.891 \scriptscriptstyle \pm \scriptstyle 6e\text{-}3$ & $0.964 \scriptscriptstyle \pm \scriptstyle 1e\text{-}3$ & $0.748 \scriptscriptstyle \pm \scriptstyle 3e\text{-}4$ \\
split wd / split lr & $0.694 \scriptscriptstyle \pm \scriptstyle 1e\text{-}2$ & $0.858 \scriptscriptstyle \pm \scriptstyle 2e\text{-}3$ & $0.431 \scriptscriptstyle \pm \scriptstyle 6e\text{-}3$ & $3.066 \scriptscriptstyle \pm \scriptstyle 3e\text{-}2$ & $0.468 \scriptscriptstyle \pm \scriptstyle 2e\text{-}3$ & $0.821 \scriptscriptstyle \pm \scriptstyle 2e\text{-}3$ & $5.632 \scriptscriptstyle \pm \scriptstyle 4e\text{-}2$ & $0.914 \scriptscriptstyle \pm \scriptstyle 1e\text{-}3$ & $1.878 \scriptscriptstyle \pm \scriptstyle 4e\text{-}3$ & $0.966 \scriptscriptstyle \pm \scriptstyle 1e\text{-}3$ & $0.748 \scriptscriptstyle \pm \scriptstyle 3e\text{-}4$ \\
\bottomrule
\end{tabular}
\end{adjustbox}}
    \label{tab:split_share_lr_wd}
\end{table}

\section{Early-stopping criterions}

Here we demonstrate that early stopping the pretraining by the value of the pretraining objective on the hold-out validation set is comparable to the early stopping by the downstream metric on the hold-out validation set after finetuning. See \autoref{tab:early_stop} for the results. This is an important practical observation, as computing pretraining objective is much faster than the full finetuning of the model, especially on large scale datasets.

\begin{table}[h!]
    \centering
    \caption{Results for single models with MLP {\footnotesize mask + target} pretraining}
    \begin{tabular}{lcccccccc}
\toprule
{} & GE \textuparrow & CH \textuparrow & CA \textdownarrow & HO \textdownarrow & OT \textdownarrow & HI \textuparrow & FB \textdownarrow & AD \textuparrow \\
\midrule
finetune early stop & $0.683$ & $0.857$ & $0.434$ & $3.056$ & $0.468$ & $0.819$ & $5.633$ & $0.914$ \\
 pretrain early stop   & $0.674$ & $0.855$ & $0.434$ & $3.031$ & $0.469$ & $0.818$ & $5.738$ & $0.914$ \\
\bottomrule
\end{tabular}

    \label{tab:early_stop}
\end{table}

\section{Extended Tables With Experimental Results}
The scores with standard deviations for single models and ensembles are provided in \ref{tab:abalation_all_results_single} and \ref{tab:abalation_all_results_ensemble} respectively.

\begin{table}[h!]
    \centering
    \caption{Extended results for single models}
    {\small 
\begin{adjustbox}{angle=90, center}
\begin{tabular}{lccccccccccc}
\toprule
{} & GE \textuparrow & CH \textuparrow & CA \textdownarrow & HO \textdownarrow & OT \textdownarrow & HI \textuparrow & FB \textdownarrow & AD \textuparrow & WE \textdownarrow & CO \textuparrow & MI \textdownarrow \\
\midrule
CatBoost        & $0.683 \scriptscriptstyle \pm \scriptstyle 4.7e\text{-}3$ & $0.864 \scriptscriptstyle \pm \scriptstyle 8.1e\text{-}4$ & $0.433 \scriptscriptstyle \pm \scriptstyle 1.7e\text{-}3$ & $3.115 \scriptscriptstyle \pm \scriptstyle 1.8e\text{-}2$ & $0.457 \scriptscriptstyle \pm \scriptstyle 1.3e\text{-}3$ & $0.806 \scriptscriptstyle \pm \scriptstyle 3.4e\text{-}4$ & $5.324 \scriptscriptstyle \pm \scriptstyle 4.0e\text{-}2$ & $0.927 \scriptscriptstyle \pm \scriptstyle 3.1e\text{-}4$ & $1.837 \scriptscriptstyle \pm \scriptstyle 2.1e\text{-}3$ & $0.966 \scriptscriptstyle \pm \scriptstyle 3.2e\text{-}4$ & $0.743 \scriptscriptstyle \pm \scriptstyle 3.0e\text{-}4$ \\
XGBoost         & $0.678 \scriptscriptstyle \pm \scriptstyle 4.8e\text{-}3$ & $0.858 \scriptscriptstyle \pm \scriptstyle 2.3e\text{-}3$ & $0.436 \scriptscriptstyle \pm \scriptstyle 2.5e\text{-}3$ & $3.160 \scriptscriptstyle \pm \scriptstyle 6.9e\text{-}3$ & $0.457 \scriptscriptstyle \pm \scriptstyle 6.0e\text{-}3$ & $0.804 \scriptscriptstyle \pm \scriptstyle 1.5e\text{-}3$ & $5.383 \scriptscriptstyle \pm \scriptstyle 2.8e\text{-}2$ & $0.927 \scriptscriptstyle \pm \scriptstyle 7.0e\text{-}4$ & $1.802 \scriptscriptstyle \pm \scriptstyle 2.0e\text{-}3$ & $0.969 \scriptscriptstyle \pm \scriptstyle 6.1e\text{-}4$ & $0.742 \scriptscriptstyle \pm \scriptstyle 1.5e\text{-}4$ \\
\midrule
\multicolumn{12}{c}{MLP} \\
\midrule
no pretraining             & $0.635 \scriptscriptstyle \pm \scriptstyle 1.3e\text{-}2$ & $0.849 \scriptscriptstyle \pm \scriptstyle 1.6e\text{-}3$ & $0.506 \scriptscriptstyle \pm \scriptstyle 8.6e\text{-}3$ & $3.156 \scriptscriptstyle \pm \scriptstyle 2.1e\text{-}2$ & $0.479 \scriptscriptstyle \pm \scriptstyle 1.4e\text{-}3$ & $0.801 \scriptscriptstyle \pm \scriptstyle 9.5e\text{-}4$ & $5.737 \scriptscriptstyle \pm \scriptstyle 6.1e\text{-}2$ & $0.908 \scriptscriptstyle \pm \scriptstyle 1.0e\text{-}3$ & $1.909 \scriptscriptstyle \pm \scriptstyle 4.6e\text{-}3$ & $0.963 \scriptscriptstyle \pm \scriptstyle 8.7e\text{-}4$ & $0.749 \scriptscriptstyle \pm \scriptstyle 3.6e\text{-}4$ \\
mask            & $0.691 \scriptscriptstyle \pm \scriptstyle 1.0e\text{-}2$ & $0.857 \scriptscriptstyle \pm \scriptstyle 2.5e\text{-}3$ & $0.454 \scriptscriptstyle \pm \scriptstyle 5.0e\text{-}3$ & $3.113 \scriptscriptstyle \pm \scriptstyle 4.3e\text{-}2$ & $0.472 \scriptscriptstyle \pm \scriptstyle 3.0e\text{-}3$ & $0.814 \scriptscriptstyle \pm \scriptstyle 1.7e\text{-}3$ & $5.681 \scriptscriptstyle \pm \scriptstyle 3.1e\text{-}2$ & $0.912 \scriptscriptstyle \pm \scriptstyle 8.0e\text{-}4$ & $1.883 \scriptscriptstyle \pm \scriptstyle 2.9e\text{-}3$ & $0.964 \scriptscriptstyle \pm \scriptstyle 9.3e\text{-}4$ & $0.748 \scriptscriptstyle \pm \scriptstyle 3.1e\text{-}4$ \\
rec             & $0.662 \scriptscriptstyle \pm \scriptstyle 1.0e\text{-}2$ & $0.853 \scriptscriptstyle \pm \scriptstyle 2.2e\text{-}3$ & $0.445 \scriptscriptstyle \pm \scriptstyle 4.0e\text{-}3$ & $3.044 \scriptscriptstyle \pm \scriptstyle 3.3e\text{-}2$ & $0.466 \scriptscriptstyle \pm \scriptstyle 2.1e\text{-}3$ & $0.805 \scriptscriptstyle \pm \scriptstyle 1.3e\text{-}3$ & $5.641 \scriptscriptstyle \pm \scriptstyle 3.2e\text{-}2$ & $0.910 \scriptscriptstyle \pm \scriptstyle 1.2e\text{-}3$ & $1.875 \scriptscriptstyle \pm \scriptstyle 3.4e\text{-}3$ & $0.965 \scriptscriptstyle \pm \scriptstyle 5.4e\text{-}4$ & $0.746 \scriptscriptstyle \pm \scriptstyle 2.3e\text{-}4$ \\
contrastive     & $0.672 \scriptscriptstyle \pm \scriptstyle 1.4e\text{-}2$ & $0.855 \scriptscriptstyle \pm \scriptstyle 2.0e\text{-}3$ & $0.455 \scriptscriptstyle \pm \scriptstyle 4.5e\text{-}3$ & $3.056 \scriptscriptstyle \pm \scriptstyle 5.2e\text{-}2$ & $0.469 \scriptscriptstyle \pm \scriptstyle 2.6e\text{-}3$ & $0.813 \scriptscriptstyle \pm \scriptstyle 1.3e\text{-}3$ & $5.697 \scriptscriptstyle \pm \scriptstyle 2.9e\text{-}2$ & $0.910 \scriptscriptstyle \pm \scriptstyle 1.3e\text{-}3$ & $1.881 \scriptscriptstyle \pm \scriptstyle 4.5e\text{-}3$ & $0.960 \scriptscriptstyle \pm \scriptstyle 1.2e\text{-}3$ & $0.748 \scriptscriptstyle \pm \scriptstyle 3.6e\text{-}4$ \\
sup             & $0.693 \scriptscriptstyle \pm \scriptstyle 1.1e\text{-}2$ & $0.856 \scriptscriptstyle \pm \scriptstyle 1.8e\text{-}3$ & $0.441 \scriptscriptstyle \pm \scriptstyle 5.3e\text{-}3$ & $3.077 \scriptscriptstyle \pm \scriptstyle 2.9e\text{-}2$ & $0.459 \scriptscriptstyle \pm \scriptstyle 2.1e\text{-}3$ & $0.814 \scriptscriptstyle \pm \scriptstyle 7.4e\text{-}4$ & $5.689 \scriptscriptstyle \pm \scriptstyle 2.1e\text{-}2$ & $0.914 \scriptscriptstyle \pm \scriptstyle 7.8e\text{-}4$ & $1.883 \scriptscriptstyle \pm \scriptstyle 4.6e\text{-}3$ & $0.968 \scriptscriptstyle \pm \scriptstyle 5.2e\text{-}4$ & $0.748 \scriptscriptstyle \pm \scriptstyle 3.0e\text{-}4$ \\
supcon          & $0.666 \scriptscriptstyle \pm \scriptstyle 1.4e\text{-}2$ & $0.850 \scriptscriptstyle \pm \scriptstyle 2.2e\text{-}3$ & $0.454 \scriptscriptstyle \pm \scriptstyle 4.2e\text{-}3$ & $3.108 \scriptscriptstyle \pm \scriptstyle 2.5e\text{-}2$ & $0.480 \scriptscriptstyle \pm \scriptstyle 1.9e\text{-}3$ & $0.806 \scriptscriptstyle \pm \scriptstyle 7.3e\text{-}4$ & $5.680 \scriptscriptstyle \pm \scriptstyle 2.4e\text{-}2$ & $0.911 \scriptscriptstyle \pm \scriptstyle 6.0e\text{-}4$ & $1.873 \scriptscriptstyle \pm \scriptstyle 2.4e\text{-}3$ & $0.966 \scriptscriptstyle \pm \scriptstyle 5.8e\text{-}4$ & $0.747 \scriptscriptstyle \pm \scriptstyle 3.2e\text{-}4$ \\
mask + sup      & $0.693 \scriptscriptstyle \pm \scriptstyle 8.2e\text{-}3$ & $0.857 \scriptscriptstyle \pm \scriptstyle 2.3e\text{-}3$ & $0.436 \scriptscriptstyle \pm \scriptstyle 6.9e\text{-}3$ & $3.099 \scriptscriptstyle \pm \scriptstyle 2.4e\text{-}2$ & $0.458 \scriptscriptstyle \pm \scriptstyle 1.7e\text{-}3$ & $0.817 \scriptscriptstyle \pm \scriptstyle 5.6e\text{-}4$ & $5.685 \scriptscriptstyle \pm \scriptstyle 3.6e\text{-}2$ & $0.915 \scriptscriptstyle \pm \scriptstyle 5.3e\text{-}4$ & $1.873 \scriptscriptstyle \pm \scriptstyle 5.1e\text{-}3$ & $0.967 \scriptscriptstyle \pm \scriptstyle 4.0e\text{-}4$ & $0.748 \scriptscriptstyle \pm \scriptstyle 2.8e\text{-}4$ \\
rec + sup       & $0.684 \scriptscriptstyle \pm \scriptstyle 7.7e\text{-}3$ & $0.854 \scriptscriptstyle \pm \scriptstyle 4.5e\text{-}3$ & $0.436 \scriptscriptstyle \pm \scriptstyle 4.4e\text{-}3$ & $3.012 \scriptscriptstyle \pm \scriptstyle 4.0e\text{-}2$ & $0.456 \scriptscriptstyle \pm \scriptstyle 1.9e\text{-}3$ & $0.815 \scriptscriptstyle \pm \scriptstyle 5.8e\text{-}4$ & $5.672 \scriptscriptstyle \pm \scriptstyle 3.6e\text{-}2$ & $0.911 \scriptscriptstyle \pm \scriptstyle 1.4e\text{-}3$ & $1.862 \scriptscriptstyle \pm \scriptstyle 2.8e\text{-}3$ & $0.967 \scriptscriptstyle \pm \scriptstyle 6.6e\text{-}4$ & $0.747 \scriptscriptstyle \pm \scriptstyle 4.9e\text{-}4$ \\
\midrule
mask + target   & $0.683 \scriptscriptstyle \pm \scriptstyle 1.0e\text{-}2$ & $0.857 \scriptscriptstyle \pm \scriptstyle 2.1e\text{-}3$ & $0.434 \scriptscriptstyle \pm \scriptstyle 7.2e\text{-}3$ & $3.056 \scriptscriptstyle \pm \scriptstyle 4.0e\text{-}2$ & $0.468 \scriptscriptstyle \pm \scriptstyle 1.9e\text{-}3$ & $0.819 \scriptscriptstyle \pm \scriptstyle 1.6e\text{-}3$ & $5.633 \scriptscriptstyle \pm \scriptstyle 3.7e\text{-}2$ & $0.914 \scriptscriptstyle \pm \scriptstyle 1.1e\text{-}3$ & $1.876 \scriptscriptstyle \pm \scriptstyle 4.8e\text{-}3$ & $0.965 \scriptscriptstyle \pm \scriptstyle 6.6e\text{-}4$ & $0.748 \scriptscriptstyle \pm \scriptstyle 4.5e\text{-}4$ \\
 - target sampling & $0.680 \scriptscriptstyle \pm \scriptstyle 9.7e\text{-}3$ & $0.857 \scriptscriptstyle \pm \scriptstyle 3.1e\text{-}3$ & $0.432 \scriptscriptstyle \pm \scriptstyle 4.9e\text{-}3$ & $3.019 \scriptscriptstyle \pm \scriptstyle 3.5e\text{-}2$ & $0.468 \scriptscriptstyle \pm \scriptstyle 1.9e\text{-}3$ & $0.815 \scriptscriptstyle \pm \scriptstyle 1.7e\text{-}3$ & $5.697 \scriptscriptstyle \pm \scriptstyle 3.1e\text{-}2$ & $0.912 \scriptscriptstyle \pm \scriptstyle 6.7e\text{-}4$ & $1.887 \scriptscriptstyle \pm \scriptstyle 3.1e\text{-}3$ & $0.964 \scriptscriptstyle \pm \scriptstyle 1.1e\text{-}3$ & $0.748 \scriptscriptstyle \pm \scriptstyle 3.1e\text{-}4$ \\
 \midrule
rec + target    & $0.659 \scriptscriptstyle \pm \scriptstyle 8.6e\text{-}3$ & $0.853 \scriptscriptstyle \pm \scriptstyle 3.2e\text{-}3$ & $0.454 \scriptscriptstyle \pm \scriptstyle 6.7e\text{-}3$ & $3.044 \scriptscriptstyle \pm \scriptstyle 4.9e\text{-}2$ & $0.463 \scriptscriptstyle \pm \scriptstyle 1.6e\text{-}3$ & $0.806 \scriptscriptstyle \pm \scriptstyle 1.5e\text{-}3$ & $5.636 \scriptscriptstyle \pm \scriptstyle 3.1e\text{-}2$ & $0.909 \scriptscriptstyle \pm \scriptstyle 9.0e\text{-}4$ & $1.884 \scriptscriptstyle \pm \scriptstyle 2.3e\text{-}3$ & $0.965 \scriptscriptstyle \pm \scriptstyle 8.7e\text{-}4$ & $0.745 \scriptscriptstyle \pm \scriptstyle 3.9e\text{-}4$ \\
 - target sampling & $0.641 \scriptscriptstyle \pm \scriptstyle 5.6e\text{-}3$ & $0.853 \scriptscriptstyle \pm \scriptstyle 3.4e\text{-}3$ & $0.455 \scriptscriptstyle \pm \scriptstyle 4.6e\text{-}3$ & $3.046 \scriptscriptstyle \pm \scriptstyle 2.4e\text{-}2$ & $0.463 \scriptscriptstyle \pm \scriptstyle 2.0e\text{-}3$ & $0.806 \scriptscriptstyle \pm \scriptstyle 1.3e\text{-}3$ & $5.640 \scriptscriptstyle \pm \scriptstyle 1.9e\text{-}2$ & $0.910 \scriptscriptstyle \pm \scriptstyle 1.1e\text{-}3$ & $1.877 \scriptscriptstyle \pm \scriptstyle 3.3e\text{-}3$ & $0.966 \scriptscriptstyle \pm \scriptstyle 4.4e\text{-}4$ & $0.746 \scriptscriptstyle \pm \scriptstyle 3.9e\text{-}4$ \\
 \midrule
\multicolumn{12}{c}{MLP-PLR} \\
\midrule
no pretraining       & $0.668 \scriptscriptstyle \pm \scriptstyle 1.4e\text{-}2$ & $0.858 \scriptscriptstyle \pm \scriptstyle 4.7e\text{-}3$ & $0.469 \scriptscriptstyle \pm \scriptstyle 5.2e\text{-}3$ & $3.008 \scriptscriptstyle \pm \scriptstyle 2.3e\text{-}2$ & $0.483 \scriptscriptstyle \pm \scriptstyle 1.6e\text{-}3$ & $0.809 \scriptscriptstyle \pm \scriptstyle 2.3e\text{-}3$ & $5.608 \scriptscriptstyle \pm \scriptstyle 5.6e\text{-}2$ & $0.926 \scriptscriptstyle \pm \scriptstyle 6.1e\text{-}4$ & $1.890 \scriptscriptstyle \pm \scriptstyle 5.0e\text{-}3$ & $0.969 \scriptscriptstyle \pm \scriptstyle 1.0e\text{-}3$ & $0.746 \scriptscriptstyle \pm \scriptstyle 3.7e\text{-}4$ \\
mask            & $0.685 \scriptscriptstyle \pm \scriptstyle 5.6e\text{-}3$ & $0.863 \scriptscriptstyle \pm \scriptstyle 1.8e\text{-}3$ & $0.434 \scriptscriptstyle \pm \scriptstyle 4.3e\text{-}3$ & $3.007 \scriptscriptstyle \pm \scriptstyle 4.6e\text{-}2$ & $0.477 \scriptscriptstyle \pm \scriptstyle 2.5e\text{-}3$ & $0.818 \scriptscriptstyle \pm \scriptstyle 8.4e\text{-}4$ & $5.586 \scriptscriptstyle \pm \scriptstyle 2.4e\text{-}2$ & $0.927 \scriptscriptstyle \pm \scriptstyle 5.1e\text{-}4$ & $1.911 \scriptscriptstyle \pm \scriptstyle 5.4e\text{-}3$ & $0.970 \scriptscriptstyle \pm \scriptstyle 5.1e\text{-}4$ & $0.748 \scriptscriptstyle \pm \scriptstyle 3.9e\text{-}4$ \\
rec             & $0.667 \scriptscriptstyle \pm \scriptstyle 7.2e\text{-}3$ & $0.852 \scriptscriptstyle \pm \scriptstyle 7.2e\text{-}3$ & $0.439 \scriptscriptstyle \pm \scriptstyle 5.2e\text{-}3$ & $3.031 \scriptscriptstyle \pm \scriptstyle 3.8e\text{-}2$ & $0.472 \scriptscriptstyle \pm \scriptstyle 3.0e\text{-}3$ & $0.808 \scriptscriptstyle \pm \scriptstyle 1.2e\text{-}3$ & $5.571 \scriptscriptstyle \pm \scriptstyle 1.2e\text{-}1$ & $0.926 \scriptscriptstyle \pm \scriptstyle 6.4e\text{-}4$ & $1.877 \scriptscriptstyle \pm \scriptstyle 4.0e\text{-}3$ & $0.971 \scriptscriptstyle \pm \scriptstyle 4.8e\text{-}4$ & $0.745 \scriptscriptstyle \pm \scriptstyle 3.6e\text{-}4$ \\
sup             & $0.710 \scriptscriptstyle \pm \scriptstyle 4.6e\text{-}3$ & $0.859 \scriptscriptstyle \pm \scriptstyle 4.1e\text{-}3$ & $0.433 \scriptscriptstyle \pm \scriptstyle 3.6e\text{-}3$ & $3.136 \scriptscriptstyle \pm \scriptstyle 8.1e\text{-}2$ & $0.479 \scriptscriptstyle \pm \scriptstyle 1.9e\text{-}3$ & $0.811 \scriptscriptstyle \pm \scriptstyle 1.0e\text{-}3$ & $5.521 \scriptscriptstyle \pm \scriptstyle 4.6e\text{-}2$ & $0.924 \scriptscriptstyle \pm \scriptstyle 1.5e\text{-}3$ & $1.873 \scriptscriptstyle \pm \scriptstyle 2.0e\text{-}3$ & $0.971 \scriptscriptstyle \pm \scriptstyle 4.5e\text{-}4$ & $0.748 \scriptscriptstyle \pm \scriptstyle 8.0e\text{-}4$ \\
mask + sup      & $0.711 \scriptscriptstyle \pm \scriptstyle 7.1e\text{-}3$ & $0.866 \scriptscriptstyle \pm \scriptstyle 2.0e\text{-}3$ & $0.441 \scriptscriptstyle \pm \scriptstyle 4.9e\text{-}3$ & $3.129 \scriptscriptstyle \pm \scriptstyle 4.1e\text{-}2$ & $0.480 \scriptscriptstyle \pm \scriptstyle 1.8e\text{-}3$ & $0.813 \scriptscriptstyle \pm \scriptstyle 8.2e\text{-}4$ & $5.480 \scriptscriptstyle \pm \scriptstyle 4.6e\text{-}2$ & $0.925 \scriptscriptstyle \pm \scriptstyle 1.0e\text{-}3$ & $1.875 \scriptscriptstyle \pm \scriptstyle 2.2e\text{-}3$ & $0.969 \scriptscriptstyle \pm \scriptstyle 6.0e\text{-}4$ & $0.745 \scriptscriptstyle \pm \scriptstyle 2.4e\text{-}4$ \\
rec + sup       & $0.709 \scriptscriptstyle \pm \scriptstyle 5.1e\text{-}3$ & $0.858 \scriptscriptstyle \pm \scriptstyle 1.9e\text{-}3$ & $0.433 \scriptscriptstyle \pm \scriptstyle 2.7e\text{-}3$ & $3.059 \scriptscriptstyle \pm \scriptstyle 3.6e\text{-}2$ & $0.465 \scriptscriptstyle \pm \scriptstyle 2.2e\text{-}3$ & $0.807 \scriptscriptstyle \pm \scriptstyle 6.2e\text{-}4$ & $5.571 \scriptscriptstyle \pm \scriptstyle 1.2e\text{-}1$ & $0.927 \scriptscriptstyle \pm \scriptstyle 5.8e\text{-}4$ & $1.865 \scriptscriptstyle \pm \scriptstyle 3.1e\text{-}3$ & $0.971 \scriptscriptstyle \pm \scriptstyle 4.4e\text{-}4$ & $0.745 \scriptscriptstyle \pm \scriptstyle 2.4e\text{-}4$ \\
 \midrule
mask + target   & $0.694 \scriptscriptstyle \pm \scriptstyle 9.1e\text{-}3$ & $0.862 \scriptscriptstyle \pm \scriptstyle 1.7e\text{-}3$ & $0.425 \scriptscriptstyle \pm \scriptstyle 4.2e\text{-}3$ & $3.023 \scriptscriptstyle \pm \scriptstyle 4.3e\text{-}2$ & $0.474 \scriptscriptstyle \pm \scriptstyle 2.0e\text{-}3$ & $0.821 \scriptscriptstyle \pm \scriptstyle 1.1e\text{-}3$ & $5.537 \scriptscriptstyle \pm \scriptstyle 3.4e\text{-}2$ & $0.929 \scriptscriptstyle \pm \scriptstyle 3.3e\text{-}4$ & $1.911 \scriptscriptstyle \pm \scriptstyle 6.2e\text{-}3$ & $0.969 \scriptscriptstyle \pm \scriptstyle 5.8e\text{-}4$ & $0.749 \scriptscriptstyle \pm \scriptstyle 1.2e\text{-}3$ \\
 - target sampling & $0.690 \scriptscriptstyle \pm \scriptstyle 9.6e\text{-}3$ & $0.864 \scriptscriptstyle \pm \scriptstyle 2.8e\text{-}3$ & $0.421 \scriptscriptstyle \pm \scriptstyle 4.5e\text{-}3$ & $2.971 \scriptscriptstyle \pm \scriptstyle 4.1e\text{-}2$ & $0.479 \scriptscriptstyle \pm \scriptstyle 2.0e\text{-}3$ & $0.821 \scriptscriptstyle \pm \scriptstyle 1.0e\text{-}3$ & $5.440 \scriptscriptstyle \pm \scriptstyle 8.1e\text{-}2$ & $0.928 \scriptscriptstyle \pm \scriptstyle 5.6e\text{-}4$ & $1.906 \scriptscriptstyle \pm \scriptstyle 5.5e\text{-}3$ & $0.970 \scriptscriptstyle \pm \scriptstyle 3.9e\text{-}4$ & $0.748 \scriptscriptstyle \pm \scriptstyle 5.3e\text{-}4$ \\
  \midrule
rec + target    & $0.688 \scriptscriptstyle \pm \scriptstyle 8.2e\text{-}3$ & $0.860 \scriptscriptstyle \pm \scriptstyle 1.4e\text{-}3$ & $0.445 \scriptscriptstyle \pm \scriptstyle 2.7e\text{-}3$ & $3.064 \scriptscriptstyle \pm \scriptstyle 3.4e\text{-}2$ & $0.475 \scriptscriptstyle \pm \scriptstyle 1.9e\text{-}3$ & $0.812 \scriptscriptstyle \pm \scriptstyle 1.1e\text{-}3$ & $5.507 \scriptscriptstyle \pm \scriptstyle 1.0e\text{-}1$ & $0.927 \scriptscriptstyle \pm \scriptstyle 3.9e\text{-}4$ & $1.887 \scriptscriptstyle \pm \scriptstyle 2.4e\text{-}3$ & $0.971 \scriptscriptstyle \pm \scriptstyle 3.9e\text{-}4$ & $0.748 \scriptscriptstyle \pm \scriptstyle 7.2e\text{-}4$ \\
 - target sampling & $0.687 \scriptscriptstyle \pm \scriptstyle 7.9e\text{-}3$ & $0.855 \scriptscriptstyle \pm \scriptstyle 4.8e\text{-}3$ & $0.453 \scriptscriptstyle \pm \scriptstyle 6.4e\text{-}3$ & $3.008 \scriptscriptstyle \pm \scriptstyle 2.7e\text{-}2$ & $0.471 \scriptscriptstyle \pm \scriptstyle 2.1e\text{-}3$ & $0.812 \scriptscriptstyle \pm \scriptstyle 5.5e\text{-}4$ & $5.592 \scriptscriptstyle \pm \scriptstyle 9.8e\text{-}2$ & $0.927 \scriptscriptstyle \pm \scriptstyle 3.4e\text{-}4$ & $1.876 \scriptscriptstyle \pm \scriptstyle 3.1e\text{-}3$ & $0.970 \scriptscriptstyle \pm \scriptstyle 4.4e\text{-}4$ & $0.745 \scriptscriptstyle \pm \scriptstyle 3.5e\text{-}4$ \\
  \midrule
\multicolumn{12}{c}{MLP-T-LR} \\
\midrule
no pretraining        & $0.634 \scriptscriptstyle \pm \scriptstyle 6.9e\text{-}3$ & $0.866 \scriptscriptstyle \pm \scriptstyle 1.3e\text{-}3$ & $0.444 \scriptscriptstyle \pm \scriptstyle 1.8e\text{-}3$ & $3.113 \scriptscriptstyle \pm \scriptstyle 4.5e\text{-}2$ & $0.482 \scriptscriptstyle \pm \scriptstyle 1.7e\text{-}3$ & $0.805 \scriptscriptstyle \pm \scriptstyle 9.3e\text{-}4$ & $5.520 \scriptscriptstyle \pm \scriptstyle 3.6e\text{-}2$ & $0.925 \scriptscriptstyle \pm \scriptstyle 6.8e\text{-}4$ & $1.897 \scriptscriptstyle \pm \scriptstyle 4.5e\text{-}3$ & $0.968 \scriptscriptstyle \pm \scriptstyle 5.0e\text{-}4$ & $0.749 \scriptscriptstyle \pm \scriptstyle 5.2e\text{-}4$ \\
mask            & $0.654 \scriptscriptstyle \pm \scriptstyle 6.4e\text{-}3$ & $0.868 \scriptscriptstyle \pm \scriptstyle 1.0e\text{-}3$ & $0.424 \scriptscriptstyle \pm \scriptstyle 2.4e\text{-}3$ & $3.045 \scriptscriptstyle \pm \scriptstyle 3.7e\text{-}2$ & $0.472 \scriptscriptstyle \pm \scriptstyle 2.5e\text{-}3$ & $0.818 \scriptscriptstyle \pm \scriptstyle 1.8e\text{-}3$ & $5.544 \scriptscriptstyle \pm \scriptstyle 3.5e\text{-}2$ & $0.926 \scriptscriptstyle \pm \scriptstyle 7.1e\text{-}4$ & $1.916 \scriptscriptstyle \pm \scriptstyle 3.1e\text{-}3$ & $0.969 \scriptscriptstyle \pm \scriptstyle 4.6e\text{-}4$ & $0.748 \scriptscriptstyle \pm \scriptstyle 3.5e\text{-}4$ \\
rec             & $0.652 \scriptscriptstyle \pm \scriptstyle 7.4e\text{-}3$ & $0.857 \scriptscriptstyle \pm \scriptstyle 4.4e\text{-}3$ & $0.424 \scriptscriptstyle \pm \scriptstyle 3.1e\text{-}3$ & $3.109 \scriptscriptstyle \pm \scriptstyle 3.7e\text{-}2$ & $0.472 \scriptscriptstyle \pm \scriptstyle 2.0e\text{-}3$ & $0.808 \scriptscriptstyle \pm \scriptstyle 1.0e\text{-}3$ & $5.363 \scriptscriptstyle \pm \scriptstyle 6.6e\text{-}2$ & $0.924 \scriptscriptstyle \pm \scriptstyle 1.8e\text{-}4$ & $1.861 \scriptscriptstyle \pm \scriptstyle 3.9e\text{-}3$ & $0.969 \scriptscriptstyle \pm \scriptstyle 7.3e\text{-}4$ & $0.746 \scriptscriptstyle \pm \scriptstyle 4.8e\text{-}4$ \\
sup             & $0.682 \scriptscriptstyle \pm \scriptstyle 5.1e\text{-}3$ & $0.860 \scriptscriptstyle \pm \scriptstyle 4.1e\text{-}3$ & $0.430 \scriptscriptstyle \pm \scriptstyle 1.9e\text{-}3$ & $3.135 \scriptscriptstyle \pm \scriptstyle 1.7e\text{-}2$ & $0.471 \scriptscriptstyle \pm \scriptstyle 1.2e\text{-}3$ & $0.807 \scriptscriptstyle \pm \scriptstyle 6.2e\text{-}4$ & $5.525 \scriptscriptstyle \pm \scriptstyle 2.3e\text{-}2$ & $0.927 \scriptscriptstyle \pm \scriptstyle 3.8e\text{-}4$ & $1.893 \scriptscriptstyle \pm \scriptstyle 2.4e\text{-}3$ & $0.971 \scriptscriptstyle \pm \scriptstyle 5.8e\text{-}4$ & $0.747 \scriptscriptstyle \pm \scriptstyle 3.1e\text{-}4$ \\
mask + sup      & $0.676 \scriptscriptstyle \pm \scriptstyle 7.7e\text{-}3$ & $0.858 \scriptscriptstyle \pm \scriptstyle 7.9e\text{-}3$ & $0.429 \scriptscriptstyle \pm \scriptstyle 1.8e\text{-}3$ & $3.199 \scriptscriptstyle \pm \scriptstyle 2.4e\text{-}2$ & $0.468 \scriptscriptstyle \pm \scriptstyle 9.9e\text{-}4$ & $0.814 \scriptscriptstyle \pm \scriptstyle 8.2e\text{-}4$ & $5.510 \scriptscriptstyle \pm \scriptstyle 3.6e\text{-}2$ & $0.926 \scriptscriptstyle \pm \scriptstyle 8.7e\text{-}4$ & $1.869 \scriptscriptstyle \pm \scriptstyle 2.9e\text{-}3$ & $0.971 \scriptscriptstyle \pm \scriptstyle 3.9e\text{-}4$ & $0.748 \scriptscriptstyle \pm \scriptstyle 2.5e\text{-}4$ \\
rec + sup       & $0.678 \scriptscriptstyle \pm \scriptstyle 6.7e\text{-}3$ & $0.865 \scriptscriptstyle \pm \scriptstyle 1.1e\text{-}3$ & $0.437 \scriptscriptstyle \pm \scriptstyle 2.0e\text{-}3$ & $3.112 \scriptscriptstyle \pm \scriptstyle 4.1e\text{-}2$ & $0.462 \scriptscriptstyle \pm \scriptstyle 2.2e\text{-}3$ & $0.807 \scriptscriptstyle \pm \scriptstyle 2.6e\text{-}3$ & $5.516 \scriptscriptstyle \pm \scriptstyle 2.2e\text{-}2$ & $0.927 \scriptscriptstyle \pm \scriptstyle 3.7e\text{-}4$ & $1.862 \scriptscriptstyle \pm \scriptstyle 3.4e\text{-}3$ & $0.970 \scriptscriptstyle \pm \scriptstyle 5.0e\text{-}4$ & $0.748 \scriptscriptstyle \pm \scriptstyle 4.2e\text{-}4$ \\
\midrule
mask + target   & $0.649 \scriptscriptstyle \pm \scriptstyle 7.3e\text{-}3$ & $0.865 \scriptscriptstyle \pm \scriptstyle 1.7e\text{-}3$ & $0.421 \scriptscriptstyle \pm \scriptstyle 3.9e\text{-}3$ & $3.058 \scriptscriptstyle \pm \scriptstyle 4.3e\text{-}2$ & $0.474 \scriptscriptstyle \pm \scriptstyle 1.9e\text{-}3$ & $0.820 \scriptscriptstyle \pm \scriptstyle 1.2e\text{-}3$ & $5.644 \scriptscriptstyle \pm \scriptstyle 4.6e\text{-}2$ & $0.929 \scriptscriptstyle \pm \scriptstyle 3.5e\text{-}4$ & $1.924 \scriptscriptstyle \pm \scriptstyle 7.6e\text{-}3$ & $0.969 \scriptscriptstyle \pm \scriptstyle 4.8e\text{-}4$ & $0.749 \scriptscriptstyle \pm \scriptstyle 5.1e\text{-}4$ \\
 - target sampling & $0.649 \scriptscriptstyle \pm \scriptstyle 8.3e\text{-}3$ & $0.861 \scriptscriptstyle \pm \scriptstyle 3.2e\text{-}3$ & $0.417 \scriptscriptstyle \pm \scriptstyle 4.9e\text{-}3$ & $3.050 \scriptscriptstyle \pm \scriptstyle 3.3e\text{-}2$ & $0.476 \scriptscriptstyle \pm \scriptstyle 1.7e\text{-}3$ & $0.819 \scriptscriptstyle \pm \scriptstyle 1.6e\text{-}3$ & $5.492 \scriptscriptstyle \pm \scriptstyle 3.2e\text{-}2$ & $0.928 \scriptscriptstyle \pm \scriptstyle 5.6e\text{-}4$ & $1.874 \scriptscriptstyle \pm \scriptstyle 3.7e\text{-}3$ & $0.969 \scriptscriptstyle \pm \scriptstyle 4.0e\text{-}4$ & $0.749 \scriptscriptstyle \pm \scriptstyle 1.4e\text{-}3$ \\
 \midrule
rec + target    & $0.668 \scriptscriptstyle \pm \scriptstyle 7.0e\text{-}3$ & $0.864 \scriptscriptstyle \pm \scriptstyle 1.7e\text{-}3$ & $0.440 \scriptscriptstyle \pm \scriptstyle 3.5e\text{-}3$ & $3.113 \scriptscriptstyle \pm \scriptstyle 3.5e\text{-}2$ & $0.473 \scriptscriptstyle \pm \scriptstyle 4.3e\text{-}3$ & $0.806 \scriptscriptstyle \pm \scriptstyle 1.0e\text{-}3$ & $5.493 \scriptscriptstyle \pm \scriptstyle 4.4e\text{-}2$ & $0.927 \scriptscriptstyle \pm \scriptstyle 5.6e\text{-}4$ & $1.862 \scriptscriptstyle \pm \scriptstyle 3.5e\text{-}3$ & $0.969 \scriptscriptstyle \pm \scriptstyle 6.5e\text{-}4$ & $0.746 \scriptscriptstyle \pm \scriptstyle 3.4e\text{-}4$ \\
 - target sampling & $0.667 \scriptscriptstyle \pm \scriptstyle 1.0e\text{-}2$ & $0.859 \scriptscriptstyle \pm \scriptstyle 2.7e\text{-}3$ & $0.432 \scriptscriptstyle \pm \scriptstyle 3.3e\text{-}3$ & $3.104 \scriptscriptstyle \pm \scriptstyle 2.9e\text{-}2$ & $0.470 \scriptscriptstyle \pm \scriptstyle 2.2e\text{-}3$ & $0.806 \scriptscriptstyle \pm \scriptstyle 1.5e\text{-}3$ & $5.391 \scriptscriptstyle \pm \scriptstyle 3.9e\text{-}2$ & $0.927 \scriptscriptstyle \pm \scriptstyle 4.1e\text{-}4$ & $1.867 \scriptscriptstyle \pm \scriptstyle 3.8e\text{-}3$ & $0.968 \scriptscriptstyle \pm \scriptstyle 7.5e\text{-}4$ & $0.746 \scriptscriptstyle \pm \scriptstyle 3.1e\text{-}4$ \\
\bottomrule
\end{tabular}
\end{adjustbox}}
    \label{tab:abalation_all_results_single}
\end{table}

\begin{table}[h!]
    \centering
    \caption{Extended results for ensemble models}
   {\small \begin{adjustbox}{angle=90,center}
\begin{tabular}{lccccccccccc}
\toprule
{} & GE \textuparrow & CH \textuparrow & CA \textdownarrow & HO \textdownarrow & OT \textdownarrow & HI \textuparrow & FB \textdownarrow & AD \textuparrow & WE \textdownarrow & CO \textuparrow & MI \textdownarrow \\
\midrule
CatBoost        & $0.692 \scriptscriptstyle \pm \scriptstyle 1.8e\text{-}3$ & $0.864 \scriptscriptstyle \pm \scriptstyle 6.8e\text{-}5$ & $0.430 \scriptscriptstyle \pm \scriptstyle 1.1e\text{-}3$ & $3.093 \scriptscriptstyle \pm \scriptstyle 5.1e\text{-}3$ & $0.450 \scriptscriptstyle \pm \scriptstyle 3.5e\text{-}4$ & $0.807 \scriptscriptstyle \pm \scriptstyle 7.5e\text{-}5$ & $5.226 \scriptscriptstyle \pm \scriptstyle 1.2e\text{-}2$ & $0.928 \scriptscriptstyle \pm \scriptstyle 1.3e\text{-}4$ & $1.801 \scriptscriptstyle \pm \scriptstyle 1.2e\text{-}3$ & $0.967 \scriptscriptstyle \pm \scriptstyle 1.3e\text{-}4$ & $0.741 \scriptscriptstyle \pm \scriptstyle 1.4e\text{-}4$ \\
XGBoost         & $0.683 \scriptscriptstyle \pm \scriptstyle 1.3e\text{-}3$ & $0.860 \scriptscriptstyle \pm \scriptstyle 4.3e\text{-}4$ & $0.434 \scriptscriptstyle \pm \scriptstyle 7.0e\text{-}4$ & $3.152 \scriptscriptstyle \pm \scriptstyle 1.2e\text{-}3$ & $0.454 \scriptscriptstyle \pm \scriptstyle 2.5e\text{-}3$ & $0.805 \scriptscriptstyle \pm \scriptstyle 8.3e\text{-}4$ & $5.338 \scriptscriptstyle \pm \scriptstyle 1.8e\text{-}2$ & $0.927 \scriptscriptstyle \pm \scriptstyle 2.1e\text{-}4$ & $1.782 \scriptscriptstyle \pm \scriptstyle 4.9e\text{-}4$ & $0.969 \scriptscriptstyle \pm \scriptstyle 8.8e\text{-}5$ & $0.742 \scriptscriptstyle \pm \scriptstyle 5.3e\text{-}5$ \\
\midrule
\multicolumn{12}{c}{MLP} \\
\midrule
no pretraining             & $0.656 \scriptscriptstyle \pm \scriptstyle 5.9e\text{-}3$ & $0.852 \scriptscriptstyle \pm \scriptstyle 5.2e\text{-}4$ & $0.482 \scriptscriptstyle \pm \scriptstyle 2.9e\text{-}3$ & $3.055 \scriptscriptstyle \pm \scriptstyle 8.4e\text{-}3$ & $0.467 \scriptscriptstyle \pm \scriptstyle 2.0e\text{-}3$ & $0.805 \scriptscriptstyle \pm \scriptstyle 2.9e\text{-}4$ & $5.666 \scriptscriptstyle \pm \scriptstyle 2.6e\text{-}3$ & $0.910 \scriptscriptstyle \pm \scriptstyle 2.7e\text{-}4$ & $1.850 \scriptscriptstyle \pm \scriptstyle 1.0e\text{-}3$ & $0.968 \scriptscriptstyle \pm \scriptstyle 2.5e\text{-}4$ & $0.747 \scriptscriptstyle \pm \scriptstyle 8.6e\text{-}5$ \\
mask            & $0.722 \scriptscriptstyle \pm \scriptstyle 1.6e\text{-}3$ & $0.859 \scriptscriptstyle \pm \scriptstyle 6.4e\text{-}4$ & $0.437 \scriptscriptstyle \pm \scriptstyle 8.1e\text{-}4$ & $3.026 \scriptscriptstyle \pm \scriptstyle 6.3e\text{-}3$ & $0.451 \scriptscriptstyle \pm \scriptstyle 1.6e\text{-}3$ & $0.824 \scriptscriptstyle \pm \scriptstyle 6.5e\text{-}4$ & $5.578 \scriptscriptstyle \pm \scriptstyle 6.8e\text{-}3$ & $0.913 \scriptscriptstyle \pm \scriptstyle 3.0e\text{-}4$ & $1.828 \scriptscriptstyle \pm \scriptstyle 1.0e\text{-}3$ & $0.967 \scriptscriptstyle \pm \scriptstyle 1.1e\text{-}4$ & $0.746 \scriptscriptstyle \pm \scriptstyle 7.7e\text{-}5$ \\
rec             & $0.679 \scriptscriptstyle \pm \scriptstyle 2.0e\text{-}3$ & $0.856 \scriptscriptstyle \pm \scriptstyle 2.8e\text{-}4$ & $0.424 \scriptscriptstyle \pm \scriptstyle 1.0e\text{-}4$ & $2.967 \scriptscriptstyle \pm \scriptstyle 6.9e\text{-}3$ & $0.453 \scriptscriptstyle \pm \scriptstyle 1.2e\text{-}3$ & $0.812 \scriptscriptstyle \pm \scriptstyle 2.5e\text{-}4$ & $5.571 \scriptscriptstyle \pm \scriptstyle 1.4e\text{-}2$ & $0.912 \scriptscriptstyle \pm \scriptstyle 7.5e\text{-}5$ & $1.811 \scriptscriptstyle \pm \scriptstyle 1.4e\text{-}3$ & $0.972 \scriptscriptstyle \pm \scriptstyle 2.0e\text{-}4$ & $0.744 \scriptscriptstyle \pm \scriptstyle 7.8e\text{-}5$ \\
contrastive     & $0.708 \scriptscriptstyle \pm \scriptstyle 4.4e\text{-}3$ & $0.857 \scriptscriptstyle \pm \scriptstyle 8.8e\text{-}4$ & $0.434 \scriptscriptstyle \pm \scriptstyle 4.0e\text{-}3$ & $2.952 \scriptscriptstyle \pm \scriptstyle 4.4e\text{-}3$ & $0.451 \scriptscriptstyle \pm \scriptstyle 6.1e\text{-}4$ & $0.820 \scriptscriptstyle \pm \scriptstyle 4.5e\text{-}5$ & $5.634 \scriptscriptstyle \pm \scriptstyle 1.4e\text{-}2$ & $0.912 \scriptscriptstyle \pm \scriptstyle 1.3e\text{-}4$ & $1.804 \scriptscriptstyle \pm \scriptstyle 2.8e\text{-}3$ & $0.964 \scriptscriptstyle \pm \scriptstyle 2.1e\text{-}4$ & $0.746 \scriptscriptstyle \pm \scriptstyle 1.7e\text{-}4$ \\
sup             & $0.717 \scriptscriptstyle \pm \scriptstyle 2.2e\text{-}3$ & $0.857 \scriptscriptstyle \pm \scriptstyle 7.4e\text{-}4$ & $0.424 \scriptscriptstyle \pm \scriptstyle 9.4e\text{-}4$ & $3.022 \scriptscriptstyle \pm \scriptstyle 1.9e\text{-}2$ & $0.443 \scriptscriptstyle \pm \scriptstyle 1.9e\text{-}3$ & $0.816 \scriptscriptstyle \pm \scriptstyle 1.4e\text{-}4$ & $5.602 \scriptscriptstyle \pm \scriptstyle 6.5e\text{-}3$ & $0.916 \scriptscriptstyle \pm \scriptstyle 7.0e\text{-}5$ & $1.828 \scriptscriptstyle \pm \scriptstyle 4.7e\text{-}3$ & $0.973 \scriptscriptstyle \pm \scriptstyle 3.6e\text{-}4$ & $0.746 \scriptscriptstyle \pm \scriptstyle 2.1e\text{-}4$ \\
supcon          & $0.686 \scriptscriptstyle \pm \scriptstyle 5.2e\text{-}3$ & $0.851 \scriptscriptstyle \pm \scriptstyle 8.4e\text{-}4$ & $0.434 \scriptscriptstyle \pm \scriptstyle 3.0e\text{-}3$ & $3.014 \scriptscriptstyle \pm \scriptstyle 6.0e\text{-}3$ & $0.465 \scriptscriptstyle \pm \scriptstyle 1.3e\text{-}3$ & $0.809 \scriptscriptstyle \pm \scriptstyle 1.3e\text{-}4$ & $5.579 \scriptscriptstyle \pm \scriptstyle 3.4e\text{-}3$ & $0.912 \scriptscriptstyle \pm \scriptstyle 3.2e\text{-}4$ & $1.827 \scriptscriptstyle \pm \scriptstyle 9.4e\text{-}4$ & $0.970 \scriptscriptstyle \pm \scriptstyle 2.1e\text{-}4$ & $0.745 \scriptscriptstyle \pm \scriptstyle 5.3e\text{-}5$ \\
mask + sup      & $0.716 \scriptscriptstyle \pm \scriptstyle 5.7e\text{-}3$ & $0.859 \scriptscriptstyle \pm \scriptstyle 1.1e\text{-}3$ & $0.418 \scriptscriptstyle \pm \scriptstyle 2.2e\text{-}3$ & $3.066 \scriptscriptstyle \pm \scriptstyle 1.2e\text{-}2$ & $0.443 \scriptscriptstyle \pm \scriptstyle 1.5e\text{-}3$ & $0.819 \scriptscriptstyle \pm \scriptstyle 1.3e\text{-}4$ & $5.601 \scriptscriptstyle \pm \scriptstyle 4.7e\text{-}3$ & $0.916 \scriptscriptstyle \pm \scriptstyle 2.0e\text{-}4$ & $1.810 \scriptscriptstyle \pm \scriptstyle 2.9e\text{-}4$ & $0.973 \scriptscriptstyle \pm \scriptstyle 3.9e\text{-}5$ & $0.747 \scriptscriptstyle \pm \scriptstyle 1.3e\text{-}4$ \\
rec + sup       & $0.709 \scriptscriptstyle \pm \scriptstyle 3.7e\text{-}3$ & $0.859 \scriptscriptstyle \pm \scriptstyle 1.8e\text{-}3$ & $0.419 \scriptscriptstyle \pm \scriptstyle 2.1e\text{-}3$ & $2.951 \scriptscriptstyle \pm \scriptstyle 1.9e\text{-}2$ & $0.442 \scriptscriptstyle \pm \scriptstyle 8.6e\text{-}4$ & $0.817 \scriptscriptstyle \pm \scriptstyle 1.0e\text{-}4$ & $5.531 \scriptscriptstyle \pm \scriptstyle 3.0e\text{-}3$ & $0.913 \scriptscriptstyle \pm \scriptstyle 5.2e\text{-}4$ & $1.801 \scriptscriptstyle \pm \scriptstyle 4.0e\text{-}3$ & $0.973 \scriptscriptstyle \pm \scriptstyle 2.6e\text{-}5$ & $0.745 \scriptscriptstyle \pm \scriptstyle 1.1e\text{-}4$ \\
\midrule
mask + target   & $0.709 \scriptscriptstyle \pm \scriptstyle 7.3e\text{-}3$ & $0.860 \scriptscriptstyle \pm \scriptstyle 1.6e\text{-}3$ & $0.414 \scriptscriptstyle \pm \scriptstyle 1.1e\text{-}3$ & $2.949 \scriptscriptstyle \pm \scriptstyle 1.9e\text{-}2$ & $0.457 \scriptscriptstyle \pm \scriptstyle 5.9e\text{-}4$ & $0.828 \scriptscriptstyle \pm \scriptstyle 6.3e\text{-}4$ & $5.551 \scriptscriptstyle \pm \scriptstyle 7.2e\text{-}3$ & $0.916 \scriptscriptstyle \pm \scriptstyle 4.6e\text{-}4$ & $1.809 \scriptscriptstyle \pm \scriptstyle 5.3e\text{-}4$ & $0.969 \scriptscriptstyle \pm \scriptstyle 5.2e\text{-}5$ & $0.746 \scriptscriptstyle \pm \scriptstyle 1.9e\text{-}4$ \\
 - target sampling & $0.706 \scriptscriptstyle \pm \scriptstyle 4.9e\text{-}3$ & $0.860 \scriptscriptstyle \pm \scriptstyle 1.1e\text{-}3$ & $0.410 \scriptscriptstyle \pm \scriptstyle 1.5e\text{-}3$ & $2.955 \scriptscriptstyle \pm \scriptstyle 2.1e\text{-}2$ & $0.456 \scriptscriptstyle \pm \scriptstyle 3.8e\text{-}4$ & $0.822 \scriptscriptstyle \pm \scriptstyle 1.1e\text{-}3$ & $5.601 \scriptscriptstyle \pm \scriptstyle 2.0e\text{-}2$ & $0.914 \scriptscriptstyle \pm \scriptstyle 2.5e\text{-}4$ & $1.837 \scriptscriptstyle \pm \scriptstyle 1.0e\text{-}3$ & $0.968 \scriptscriptstyle \pm \scriptstyle 2.8e\text{-}4$ & $0.746 \scriptscriptstyle \pm \scriptstyle 2.1e\text{-}4$ \\
 \midrule
rec + target    & $0.677 \scriptscriptstyle \pm \scriptstyle 4.6e\text{-}3$ & $0.857 \scriptscriptstyle \pm \scriptstyle 3.9e\text{-}4$ & $0.433 \scriptscriptstyle \pm \scriptstyle 1.1e\text{-}3$ & $2.926 \scriptscriptstyle \pm \scriptstyle 2.3e\text{-}2$ & $0.448 \scriptscriptstyle \pm \scriptstyle 9.4e\text{-}4$ & $0.816 \scriptscriptstyle \pm \scriptstyle 5.3e\text{-}4$ & $5.555 \scriptscriptstyle \pm \scriptstyle 5.8e\text{-}3$ & $0.910 \scriptscriptstyle \pm \scriptstyle 2.2e\text{-}4$ & $1.825 \scriptscriptstyle \pm \scriptstyle 2.5e\text{-}3$ & $0.972 \scriptscriptstyle \pm \scriptstyle 6.1e\text{-}5$ & $0.743 \scriptscriptstyle \pm \scriptstyle 1.2e\text{-}4$ \\
 - target sampling & $0.669 \scriptscriptstyle \pm \scriptstyle 3.7e\text{-}3$ & $0.858 \scriptscriptstyle \pm \scriptstyle 1.4e\text{-}3$ & $0.435 \scriptscriptstyle \pm \scriptstyle 8.2e\text{-}4$ & $3.003 \scriptscriptstyle \pm \scriptstyle 1.1e\text{-}2$ & $0.451 \scriptscriptstyle \pm \scriptstyle 1.2e\text{-}3$ & $0.815 \scriptscriptstyle \pm \scriptstyle 7.0e\text{-}4$ & $5.577 \scriptscriptstyle \pm \scriptstyle 1.1e\text{-}2$ & $0.913 \scriptscriptstyle \pm \scriptstyle 2.4e\text{-}4$ & $1.822 \scriptscriptstyle \pm \scriptstyle 6.6e\text{-}4$ & $0.972 \scriptscriptstyle \pm \scriptstyle 2.7e\text{-}4$ & $0.744 \scriptscriptstyle \pm \scriptstyle 7.6e\text{-}5$ \\
\midrule
\multicolumn{12}{c}{MLP-PLR} \\
\midrule
no pretraining       & $0.695 \scriptscriptstyle \pm \scriptstyle 3.7e\text{-}3$ & $0.864 \scriptscriptstyle \pm \scriptstyle 7.6e\text{-}4$ & $0.454 \scriptscriptstyle \pm \scriptstyle 1.2e\text{-}3$ & $2.953 \scriptscriptstyle \pm \scriptstyle 7.4e\text{-}3$ & $0.470 \scriptscriptstyle \pm \scriptstyle 7.5e\text{-}4$ & $0.814 \scriptscriptstyle \pm \scriptstyle 7.9e\text{-}4$ & $5.324 \scriptscriptstyle \pm \scriptstyle 3.2e\text{-}2$ & $0.928 \scriptscriptstyle \pm \scriptstyle 7.7e\text{-}5$ & $1.835 \scriptscriptstyle \pm \scriptstyle 1.5e\text{-}3$ & $0.974 \scriptscriptstyle \pm \scriptstyle 2.2e\text{-}4$ & $0.744 \scriptscriptstyle \pm \scriptstyle 1.2e\text{-}4$ \\
mask            & $0.725 \scriptscriptstyle \pm \scriptstyle 4.9e\text{-}3$ & $0.865 \scriptscriptstyle \pm \scriptstyle 7.0e\text{-}4$ & $0.421 \scriptscriptstyle \pm \scriptstyle 1.7e\text{-}3$ & $2.921 \scriptscriptstyle \pm \scriptstyle 1.0e\text{-}2$ & $0.457 \scriptscriptstyle \pm \scriptstyle 8.4e\text{-}4$ & $0.827 \scriptscriptstyle \pm \scriptstyle 1.1e\text{-}4$ & $5.444 \scriptscriptstyle \pm \scriptstyle 5.4e\text{-}3$ & $0.928 \scriptscriptstyle \pm \scriptstyle 1.0e\text{-}4$ & $1.850 \scriptscriptstyle \pm \scriptstyle 3.4e\text{-}3$ & $0.974 \scriptscriptstyle \pm \scriptstyle 2.3e\text{-}4$ & $0.745 \scriptscriptstyle \pm \scriptstyle 6.5e\text{-}5$ \\
rec             & $0.698 \scriptscriptstyle \pm \scriptstyle 1.5e\text{-}3$ & $0.857 \scriptscriptstyle \pm \scriptstyle 1.5e\text{-}3$ & $0.418 \scriptscriptstyle \pm \scriptstyle 1.2e\text{-}3$ & $2.954 \scriptscriptstyle \pm \scriptstyle 8.1e\text{-}3$ & $0.454 \scriptscriptstyle \pm \scriptstyle 1.9e\text{-}3$ & $0.813 \scriptscriptstyle \pm \scriptstyle 7.8e\text{-}4$ & $5.124 \scriptscriptstyle \pm \scriptstyle 2.4e\text{-}2$ & $0.928 \scriptscriptstyle \pm \scriptstyle 2.2e\text{-}4$ & $1.818 \scriptscriptstyle \pm \scriptstyle 2.5e\text{-}3$ & $0.975 \scriptscriptstyle \pm \scriptstyle 2.9e\text{-}4$ & $0.743 \scriptscriptstyle \pm \scriptstyle 2.1e\text{-}4$ \\
sup             & $0.733 \scriptscriptstyle \pm \scriptstyle 2.2e\text{-}3$ & $0.867 \scriptscriptstyle \pm \scriptstyle 9.6e\text{-}4$ & $0.421 \scriptscriptstyle \pm \scriptstyle 1.0e\text{-}3$ & $3.054 \scriptscriptstyle \pm \scriptstyle 4.5e\text{-}2$ & $0.465 \scriptscriptstyle \pm \scriptstyle 1.3e\text{-}3$ & $0.816 \scriptscriptstyle \pm \scriptstyle 1.4e\text{-}4$ & $5.407 \scriptscriptstyle \pm \scriptstyle 3.8e\text{-}2$ & $0.926 \scriptscriptstyle \pm \scriptstyle 4.3e\text{-}4$ & $1.834 \scriptscriptstyle \pm \scriptstyle 3.3e\text{-}4$ & $0.975 \scriptscriptstyle \pm \scriptstyle 1.6e\text{-}4$ & $0.746 \scriptscriptstyle \pm \scriptstyle 2.1e\text{-}4$ \\
mask + sup      & $0.732 \scriptscriptstyle \pm \scriptstyle 2.0e\text{-}3$ & $0.869 \scriptscriptstyle \pm \scriptstyle 3.5e\text{-}4$ & $0.424 \scriptscriptstyle \pm \scriptstyle 8.9e\text{-}4$ & $3.055 \scriptscriptstyle \pm \scriptstyle 1.6e\text{-}2$ & $0.468 \scriptscriptstyle \pm \scriptstyle 7.8e\text{-}4$ & $0.817 \scriptscriptstyle \pm \scriptstyle 3.4e\text{-}4$ & $5.366 \scriptscriptstyle \pm \scriptstyle 1.1e\text{-}2$ & $0.927 \scriptscriptstyle \pm \scriptstyle 2.1e\text{-}4$ & $1.848 \scriptscriptstyle \pm \scriptstyle 1.7e\text{-}3$ & $0.974 \scriptscriptstyle \pm \scriptstyle 2.1e\text{-}4$ & $0.744 \scriptscriptstyle \pm \scriptstyle 7.9e\text{-}5$ \\
rec + sup       & $0.737 \scriptscriptstyle \pm \scriptstyle 2.0e\text{-}3$ & $0.862 \scriptscriptstyle \pm \scriptstyle 1.3e\text{-}3$ & $0.424 \scriptscriptstyle \pm \scriptstyle 9.9e\text{-}4$ & $2.964 \scriptscriptstyle \pm \scriptstyle 2.3e\text{-}3$ & $0.449 \scriptscriptstyle \pm \scriptstyle 2.3e\text{-}4$ & $0.811 \scriptscriptstyle \pm \scriptstyle 5.3e\text{-}4$ & $5.124 \scriptscriptstyle \pm \scriptstyle 2.4e\text{-}2$ & $0.929 \scriptscriptstyle \pm \scriptstyle 1.7e\text{-}4$ & $1.813 \scriptscriptstyle \pm \scriptstyle 1.9e\text{-}3$ & $0.974 \scriptscriptstyle \pm \scriptstyle 2.2e\text{-}4$ & $0.744 \scriptscriptstyle \pm \scriptstyle 7.2e\text{-}5$ \\
\midrule
mask + target   & $0.719 \scriptscriptstyle \pm \scriptstyle 3.5e\text{-}3$ & $0.866 \scriptscriptstyle \pm \scriptstyle 4.3e\text{-}4$ & $0.407 \scriptscriptstyle \pm \scriptstyle 8.2e\text{-}4$ & $2.952 \scriptscriptstyle \pm \scriptstyle 3.5e\text{-}3$ & $0.458 \scriptscriptstyle \pm \scriptstyle 4.2e\text{-}4$ & $0.828 \scriptscriptstyle \pm \scriptstyle 6.0e\text{-}4$ & $5.373 \scriptscriptstyle \pm \scriptstyle 1.8e\text{-}2$ & $0.930 \scriptscriptstyle \pm \scriptstyle 1.0e\text{-}4$ & $1.849 \scriptscriptstyle \pm \scriptstyle 2.1e\text{-}3$ & $0.973 \scriptscriptstyle \pm \scriptstyle 1.4e\text{-}4$ & $0.745 \scriptscriptstyle \pm \scriptstyle 2.5e\text{-}4$ \\
 - target sampling & $0.724 \scriptscriptstyle \pm \scriptstyle 7.0e\text{-}3$ & $0.867 \scriptscriptstyle \pm \scriptstyle 1.0e\text{-}3$ & $0.403 \scriptscriptstyle \pm \scriptstyle 1.1e\text{-}3$ & $2.877 \scriptscriptstyle \pm \scriptstyle 1.0e\text{-}2$ & $0.466 \scriptscriptstyle \pm \scriptstyle 9.4e\text{-}4$ & $0.828 \scriptscriptstyle \pm \scriptstyle 2.5e\text{-}4$ & $5.175 \scriptscriptstyle \pm \scriptstyle 1.5e\text{-}2$ & $0.930 \scriptscriptstyle \pm \scriptstyle 2.0e\text{-}4$ & $1.833 \scriptscriptstyle \pm \scriptstyle 3.6e\text{-}3$ & $0.974 \scriptscriptstyle \pm \scriptstyle 3.3e\text{-}4$ & $0.744 \scriptscriptstyle \pm \scriptstyle 2.0e\text{-}4$ \\
 \midrule
rec + target    & $0.705 \scriptscriptstyle \pm \scriptstyle 2.3e\text{-}3$ & $0.862 \scriptscriptstyle \pm \scriptstyle 2.0e\text{-}4$ & $0.431 \scriptscriptstyle \pm \scriptstyle 1.4e\text{-}3$ & $2.983 \scriptscriptstyle \pm \scriptstyle 1.2e\text{-}2$ & $0.465 \scriptscriptstyle \pm \scriptstyle 9.2e\text{-}4$ & $0.816 \scriptscriptstyle \pm \scriptstyle 1.7e\text{-}4$ & $5.096 \scriptscriptstyle \pm \scriptstyle 1.8e\text{-}2$ & $0.928 \scriptscriptstyle \pm \scriptstyle 1.9e\text{-}4$ & $1.860 \scriptscriptstyle \pm \scriptstyle 2.0e\text{-}3$ & $0.974 \scriptscriptstyle \pm \scriptstyle 3.4e\text{-}5$ & $0.745 \scriptscriptstyle \pm \scriptstyle 2.7e\text{-}4$ \\
 - target sampling & $0.712 \scriptscriptstyle \pm \scriptstyle 2.3e\text{-}3$ & $0.860 \scriptscriptstyle \pm \scriptstyle 9.3e\text{-}4$ & $0.437 \scriptscriptstyle \pm \scriptstyle 3.3e\text{-}3$ & $2.933 \scriptscriptstyle \pm \scriptstyle 2.0e\text{-}2$ & $0.450 \scriptscriptstyle \pm \scriptstyle 1.7e\text{-}3$ & $0.815 \scriptscriptstyle \pm \scriptstyle 3.5e\text{-}4$ & $5.173 \scriptscriptstyle \pm \scriptstyle 2.7e\text{-}2$ & $0.928 \scriptscriptstyle \pm \scriptstyle 5.6e\text{-}5$ & $1.811 \scriptscriptstyle \pm \scriptstyle 1.5e\text{-}3$ & $0.974 \scriptscriptstyle \pm \scriptstyle 3.6e\text{-}4$ & $0.744 \scriptscriptstyle \pm \scriptstyle 6.3e\text{-}5$ \\
\midrule
\multicolumn{12}{c}{MLP-T-LR} \\
\midrule
no pretraining       & $0.662 \scriptscriptstyle \pm \scriptstyle 7.6e\text{-}3$ & $0.868 \scriptscriptstyle \pm \scriptstyle 5.0e\text{-}4$ & $0.437 \scriptscriptstyle \pm \scriptstyle 8.2e\text{-}4$ & $3.028 \scriptscriptstyle \pm \scriptstyle 1.8e\text{-}2$ & $0.472 \scriptscriptstyle \pm \scriptstyle 4.7e\text{-}4$ & $0.808 \scriptscriptstyle \pm \scriptstyle 1.5e\text{-}4$ & $5.424 \scriptscriptstyle \pm \scriptstyle 2.2e\text{-}2$ & $0.927 \scriptscriptstyle \pm \scriptstyle 2.8e\text{-}4$ & $1.850 \scriptscriptstyle \pm \scriptstyle 9.0e\text{-}4$ & $0.972 \scriptscriptstyle \pm \scriptstyle 1.5e\text{-}4$ & $0.747 \scriptscriptstyle \pm \scriptstyle 7.6e\text{-}5$ \\
mask            & $0.679 \scriptscriptstyle \pm \scriptstyle 4.7e\text{-}3$ & $0.868 \scriptscriptstyle \pm \scriptstyle 2.3e\text{-}4$ & $0.413 \scriptscriptstyle \pm \scriptstyle 1.0e\text{-}3$ & $2.930 \scriptscriptstyle \pm \scriptstyle 1.2e\text{-}2$ & $0.450 \scriptscriptstyle \pm \scriptstyle 7.3e\text{-}4$ & $0.826 \scriptscriptstyle \pm \scriptstyle 1.3e\text{-}3$ & $5.370 \scriptscriptstyle \pm \scriptstyle 8.7e\text{-}3$ & $0.927 \scriptscriptstyle \pm \scriptstyle 3.1e\text{-}4$ & $1.836 \scriptscriptstyle \pm \scriptstyle 2.7e\text{-}3$ & $0.973 \scriptscriptstyle \pm \scriptstyle 8.8e\text{-}5$ & $0.745 \scriptscriptstyle \pm \scriptstyle 1.1e\text{-}4$ \\
rec             & $0.694 \scriptscriptstyle \pm \scriptstyle 3.7e\text{-}3$ & $0.861 \scriptscriptstyle \pm \scriptstyle 1.6e\text{-}4$ & $0.414 \scriptscriptstyle \pm \scriptstyle 1.5e\text{-}3$ & $3.035 \scriptscriptstyle \pm \scriptstyle 1.9e\text{-}2$ & $0.459 \scriptscriptstyle \pm \scriptstyle 3.4e\text{-}4$ & $0.812 \scriptscriptstyle \pm \scriptstyle 2.7e\text{-}4$ & $5.039 \scriptscriptstyle \pm \scriptstyle 1.8e\text{-}2$ & $0.925 \scriptscriptstyle \pm \scriptstyle 1.2e\text{-}4$ & $1.803 \scriptscriptstyle \pm \scriptstyle 2.3e\text{-}3$ & $0.973 \scriptscriptstyle \pm \scriptstyle 4.9e\text{-}5$ & $0.744 \scriptscriptstyle \pm \scriptstyle 4.5e\text{-}4$ \\
sup             & $0.698 \scriptscriptstyle \pm \scriptstyle 3.7e\text{-}3$ & $0.865 \scriptscriptstyle \pm \scriptstyle 7.0e\text{-}4$ & $0.424 \scriptscriptstyle \pm \scriptstyle 1.1e\text{-}3$ & $3.107 \scriptscriptstyle \pm \scriptstyle 6.8e\text{-}3$ & $0.463 \scriptscriptstyle \pm \scriptstyle 3.5e\text{-}4$ & $0.809 \scriptscriptstyle \pm \scriptstyle 2.5e\text{-}4$ & $5.442 \scriptscriptstyle \pm \scriptstyle 1.6e\text{-}2$ & $0.928 \scriptscriptstyle \pm \scriptstyle 1.4e\text{-}4$ & $1.849 \scriptscriptstyle \pm \scriptstyle 5.4e\text{-}4$ & $0.975 \scriptscriptstyle \pm \scriptstyle 2.8e\text{-}4$ & $0.746 \scriptscriptstyle \pm \scriptstyle 1.4e\text{-}5$ \\
mask + sup      & $0.698 \scriptscriptstyle \pm \scriptstyle 4.0e\text{-}3$ & $0.866 \scriptscriptstyle \pm \scriptstyle 1.1e\text{-}3$ & $0.421 \scriptscriptstyle \pm \scriptstyle 8.6e\text{-}4$ & $3.088 \scriptscriptstyle \pm \scriptstyle 5.1e\text{-}3$ & $0.460 \scriptscriptstyle \pm \scriptstyle 4.6e\text{-}4$ & $0.818 \scriptscriptstyle \pm \scriptstyle 2.8e\text{-}4$ & $5.407 \scriptscriptstyle \pm \scriptstyle 3.3e\text{-}3$ & $0.927 \scriptscriptstyle \pm \scriptstyle 5.1e\text{-}4$ & $1.824 \scriptscriptstyle \pm \scriptstyle 3.0e\text{-}3$ & $0.975 \scriptscriptstyle \pm \scriptstyle 1.4e\text{-}4$ & $0.747 \scriptscriptstyle \pm \scriptstyle 8.2e\text{-}5$ \\
rec + sup       & $0.705 \scriptscriptstyle \pm \scriptstyle 2.4e\text{-}3$ & $0.866 \scriptscriptstyle \pm \scriptstyle 4.6e\text{-}4$ & $0.425 \scriptscriptstyle \pm \scriptstyle 5.1e\text{-}4$ & $3.057 \scriptscriptstyle \pm \scriptstyle 1.0e\text{-}2$ & $0.444 \scriptscriptstyle \pm \scriptstyle 1.4e\text{-}3$ & $0.814 \scriptscriptstyle \pm \scriptstyle 6.8e\text{-}4$ & $5.422 \scriptscriptstyle \pm \scriptstyle 1.8e\text{-}3$ & $0.927 \scriptscriptstyle \pm \scriptstyle 6.6e\text{-}5$ & $1.811 \scriptscriptstyle \pm \scriptstyle 1.0e\text{-}3$ & $0.974 \scriptscriptstyle \pm \scriptstyle 1.1e\text{-}4$ & $0.746 \scriptscriptstyle \pm \scriptstyle 3.6e\text{-}4$ \\
\midrule
mask + target   & $0.673 \scriptscriptstyle \pm \scriptstyle 1.0e\text{-}3$ & $0.868 \scriptscriptstyle \pm \scriptstyle 4.6e\text{-}4$ & $0.410 \scriptscriptstyle \pm \scriptstyle 7.8e\text{-}4$ & $2.894 \scriptscriptstyle \pm \scriptstyle 1.8e\text{-}2$ & $0.460 \scriptscriptstyle \pm \scriptstyle 1.4e\text{-}3$ & $0.827 \scriptscriptstyle \pm \scriptstyle 3.4e\text{-}4$ & $5.458 \scriptscriptstyle \pm \scriptstyle 2.8e\text{-}2$ & $0.930 \scriptscriptstyle \pm \scriptstyle 1.7e\text{-}5$ & $1.849 \scriptscriptstyle \pm \scriptstyle 4.2e\text{-}3$ & $0.972 \scriptscriptstyle \pm \scriptstyle 2.5e\text{-}4$ & $0.746 \scriptscriptstyle \pm \scriptstyle 2.3e\text{-}4$ \\
 - target sampling & $0.677 \scriptscriptstyle \pm \scriptstyle 5.1e\text{-}3$ & $0.866 \scriptscriptstyle \pm \scriptstyle 1.1e\text{-}3$ & $0.397 \scriptscriptstyle \pm \scriptstyle 3.7e\text{-}4$ & $2.938 \scriptscriptstyle \pm \scriptstyle 1.8e\text{-}2$ & $0.462 \scriptscriptstyle \pm \scriptstyle 4.4e\text{-}4$ & $0.826 \scriptscriptstyle \pm \scriptstyle 1.2e\text{-}4$ & $5.384 \scriptscriptstyle \pm \scriptstyle 1.5e\text{-}2$ & $0.929 \scriptscriptstyle \pm \scriptstyle 1.6e\text{-}4$ & $1.840 \scriptscriptstyle \pm \scriptstyle 2.2e\text{-}3$ & $0.973 \scriptscriptstyle \pm \scriptstyle 1.1e\text{-}4$ & $0.747 \scriptscriptstyle \pm \scriptstyle 5.0e\text{-}4$ \\
 \midrule
rec + target    & $0.693 \scriptscriptstyle \pm \scriptstyle 2.5e\text{-}3$ & $0.866 \scriptscriptstyle \pm \scriptstyle 3.1e\text{-}4$ & $0.432 \scriptscriptstyle \pm \scriptstyle 6.1e\text{-}4$ & $3.045 \scriptscriptstyle \pm \scriptstyle 1.1e\text{-}2$ & $0.456 \scriptscriptstyle \pm \scriptstyle 2.4e\text{-}3$ & $0.812 \scriptscriptstyle \pm \scriptstyle 6.1e\text{-}4$ & $5.344 \scriptscriptstyle \pm \scriptstyle 8.1e\text{-}3$ & $0.928 \scriptscriptstyle \pm \scriptstyle 1.4e\text{-}4$ & $1.830 \scriptscriptstyle \pm \scriptstyle 2.4e\text{-}3$ & $0.972 \scriptscriptstyle \pm \scriptstyle 1.1e\text{-}4$ & $0.744 \scriptscriptstyle \pm \scriptstyle 2.1e\text{-}4$ \\
 - target sampling & $0.700 \scriptscriptstyle \pm \scriptstyle 2.5e\text{-}3$ & $0.863 \scriptscriptstyle \pm \scriptstyle 1.3e\text{-}3$ & $0.423 \scriptscriptstyle \pm \scriptstyle 9.8e\text{-}4$ & $3.029 \scriptscriptstyle \pm \scriptstyle 1.4e\text{-}2$ & $0.454 \scriptscriptstyle \pm \scriptstyle 1.8e\text{-}3$ & $0.811 \scriptscriptstyle \pm \scriptstyle 5.1e\text{-}4$ & $5.083 \scriptscriptstyle \pm \scriptstyle 4.7e\text{-}3$ & $0.928 \scriptscriptstyle \pm \scriptstyle 1.7e\text{-}4$ & $1.809 \scriptscriptstyle \pm \scriptstyle 1.0e\text{-}3$ & $0.972 \scriptscriptstyle \pm \scriptstyle 5.5e\text{-}5$ & $0.744 \scriptscriptstyle \pm \scriptstyle 3.0e\text{-}5$ \\
\bottomrule
\end{tabular}
\end{adjustbox}}
    \label{tab:abalation_all_results_ensemble}
\end{table}

\end{document}